\documentclass[lettersize,journal]{IEEEtran}
\usepackage{amsmath,amsfonts}
\usepackage{algorithmic}
\usepackage{algorithm}
\usepackage{array}
\usepackage[caption=false,font=normalsize,labelfont=sf,textfont=sf]{subfig}
\usepackage{textcomp}
\usepackage{stfloats}
\usepackage{url}
\usepackage{verbatim}
\usepackage{graphicx}
\usepackage{cite}

\usepackage{tikz, xcolor}
\usetikzlibrary{arrows.meta}
\definecolor{lime}{HTML}{A6CE39}
\definecolor{c1}{HTML}{d6ecf0}
\definecolor{metablue}{HTML}{0064E0}
\definecolor{metafg}{HTML}{1C2B33}
\definecolor{metabg}{HTML}{F1F4F7}

\usepackage{hyperref}
\hypersetup{
    colorlinks,
    citecolor=metablue,
}

\usepackage[capitalize]{cleveref}
\crefname{section}{Sec.}{Secs.}
\Crefname{section}{Section}{Sections}
\crefname{table}{Tab.}{Tabs.}

\newcommand{\minisection}[1]{\vspace{0.04in} \noindent \textbf{#1}\,}

\usepackage{dsfont}
\usepackage{pifont}

\usepackage{booktabs}
\usepackage{tabularx}
\usepackage{multirow}
\usepackage{makecell}
\usepackage{colortbl}

\usepackage[most]{tcolorbox}
\tcbset{
  myboxstyle/.style={
    enhanced,
    breakable,
    colback=gray!20,
    colframe=black!70,
    coltitle=white,
    fonttitle=\bfseries,
    fontupper=\footnotesize\itshape,
    boxrule=0.8mm,
    arc=1mm,
    boxsep=1mm,
    left=1mm,
    right=1mm,
    top=1mm,
    bottom=1mm,
    toptitle=0mm,
    bottomtitle=0mm,
    before upper={\setlength{\parindent}{0pt}\setlength{\emergencystretch}{2em}},
  }
}

\newcommand{\sArt}{SOTA~}

\definecolor{c1}{HTML}{d6ecf0}

\def\ie{\emph{i.e.}}
\def\eg{\emph{e.g.}}

\def\model{CORE}
\def\task{COR}
\def\dataset{COR125K}

\usepackage[ruled,vlined,algo2e]{algorithm2e}

\newcommand{\circlednum}[1]{%
    \tikz[baseline=(X.base)] {
        \node[draw,circle,inner sep=.3pt] (X) {#1};
    }%
}

\DeclareRobustCommand{\orcidicon}{
\begin{tikzpicture}
\draw[lime, fill=lime] (0,0)
circle[radius=0.16]
node[white]{{\fontfamily{qag}\selectfont \tiny \.{I}D}};
\end{tikzpicture}
\hspace{-2mm}
}
\foreach \x in {A, ..., Z}{%
\expandafter\xdef\csname orcid\x\endcsname{\noexpand\href{https://orcid.org/\csname orcidauthor\x\endcsname}{\noexpand\orcidicon}}
}

\begin{document}

\title{Composed Object Retrieval: Object-level Retrieval via Composed Expressions}

\author{Tong Wang\hspace{-1.5mm}\orcidA{},
        Guanyu Yang\hspace{-1.5mm}\orcidB{},
        Nian Liu\hspace{-1.5mm}\orcidC{},
        Zongyan Han\hspace{-1.5mm}\orcidD{},
        Jinxing Zhou\hspace{-1.5mm}\orcidE{},\\
        Salman Khan\hspace{-1.5mm}\orcidF{},
        and Fahad Shahbaz Khan\hspace{-1.5mm}\orcidG{}%
\thanks{Tong Wang and Guanyu Yang are with the Key Laboratory of New Generation Artificial Intelligence Technology and Its Interdisciplinary Applications, Southeast University, Ministry of Education, Jiangsu, China. (email: tongwangnj@qq.com, yang.list@seu.edu.cn)}%
\thanks{Nian Liu, Zongyan Han, Jinxing Zhou, Salman Khan, and Fahad Shahbaz Khan are with the Mohamed bin Zayed University of Artificial Intelligence (MBZUAI), Abu Dhabi, UAE. (email: liunian228@gmail.com)}%
\thanks{Corresponding author: \emph{Guanyu~Yang} and \emph{Nian Liu}. Work was done when Tong Wang was visiting MBZUAI.}
}

\markboth{IEEE Transactions on Pattern Analysis and Machine Intelligence}%
{Wang \emph{et al.}: Composed Object Retrieval}


\maketitle

\begin{abstract}
Retrieving fine-grained visual content based on user intent remains a challenge in multimodal systems. Although current Composed Image Retrieval (CIR) methods combine reference images with retrieval texts, they are constrained to image-level matching and cannot localize specific objects.
To this end, we propose \textbf{Composed Object Retrieval (\task{})}, a new object-level retrieval task that retrieves target object(s) from candidate objects in a target image and grounds the retrieved result with pixel-level masks.
Given a reference object, its mask, a target image, and a retrieval text describing the desired modification, \task{} requires models to perform composed visual-textual reasoning rather than relying on explicit category names.
This setting introduces several challenges, including fine-grained compositional matching, negative-object filtering under visually similar distractors, and flexible single- or multi-object retrieval.
We construct \textbf{\dataset{}}, the first large-scale \task{} benchmark, containing 125,541 retrieval triplets across 408 categories with base/novel splits for evaluating category-level generalization.
We also present \textbf{\model{}}, a unified end-to-end model that integrates reference region encoding, adaptive vision-text interaction, and region-level contrastive learning to align composed representations with target objects while suppressing background and distractors.
Extensive experiments demonstrate that \model{} significantly outperforms existing CIR-based pipelines and strong baselines in both base and novel categories, establishing a simple and effective foundation for fine-grained object-level multimodal retrieval.
Code will be released publicly at \url{https://github.com/wangtong627/COR}.
\end{abstract}


\section{Introduction}\label{sec}

\begin{figure}[ht]
\centering
\includegraphics[width=\linewidth]{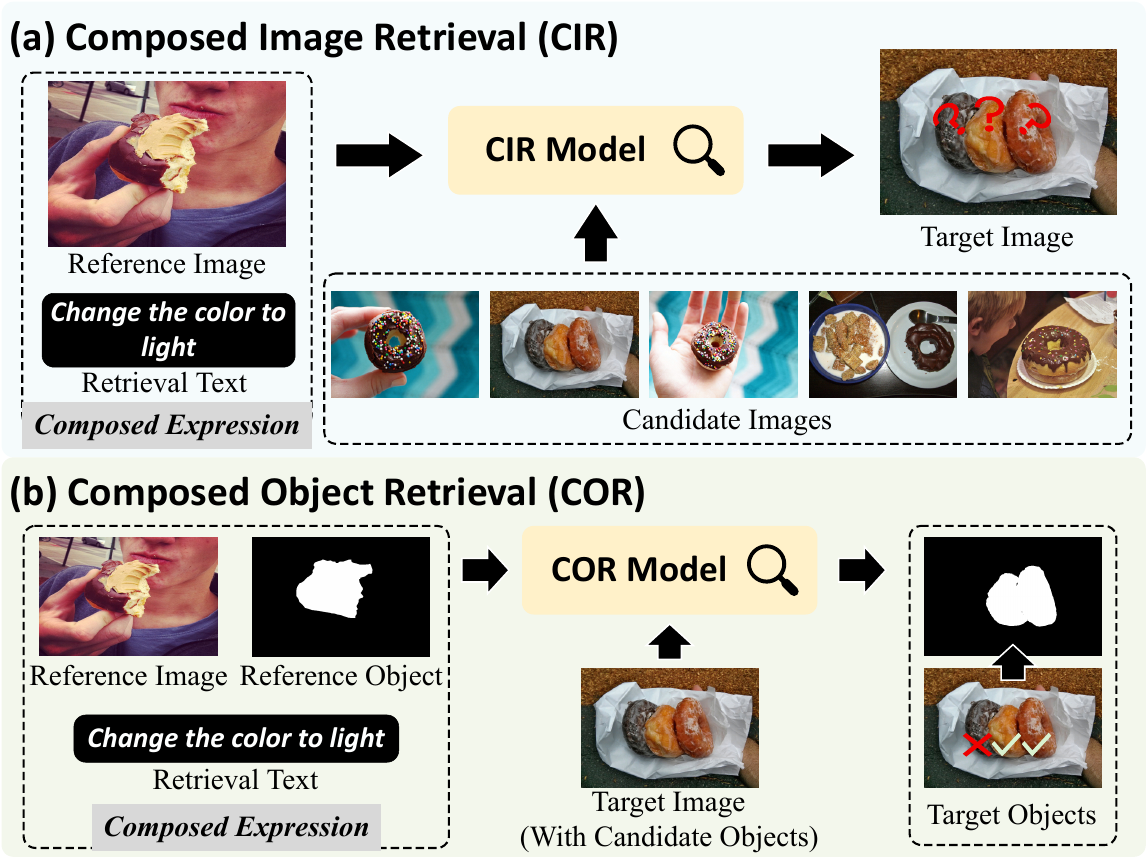}
\vspace{-8mm}
\caption{
\textbf{\task{} retrieves target objects using composed expressions.} It enables fine-grained object-level retrieval by distinguishing targets (\ie, light-colored doughnuts) from negatives (\ie, dark-colored ones). The retrieval text (\ie, \textit{``change the color to light''}) specifies attribute changes, allowing flexible retrieval based on the reference object and text without requiring explicit target object names (\ie, doughnut), thus supporting effective retrieval even when object categories are difficult to describe.
}
\vspace{-6mm}
\label{fig:intro_motivation}
\end{figure}

Image retrieval aims to match user queries with relevant visual content. Despite clear progress, traditional single-modal methods~\cite{chun2021probabilistic,dubey2021decade} often fail to capture subtle meanings and personal needs in complex multimodal scenarios.
Recently, Composed Image Retrieval (CIR)~\cite{liu2021image, baldrati2023composed, huang2024dynamic} has become an effective paradigm for retrieving target images using a reference image and a text modification (\eg, \textit{``change the color''}). By combining visual details with textual intent, CIR is useful for e-commerce~\cite{dong2023entity,chen2025prvr,ebrahimi2022heterogeneous}, social media~\cite{qi2011exploring,jafarian2022self,shivakumara2023new,mishra2024subjective}, and interactive visual search~\cite{guo2018dialog,kim2021dual,gosselin2008active,rossetto2020interactive,liang2023simple}, where users may know the desired change but cannot describe the target with a single category name or keyword.
However, CIR works at the image level, which limits fine-grained object understanding and precise localization.
As shown in \cref{fig:intro_motivation}(a), CIR retrieves whole images that may contain both matching objects (\ie, light-colored doughnuts) and non-matching ones (\ie, dark-colored doughnuts). Thus, the results can be ambiguous and often require manual filtering, reducing practical efficiency.
More importantly, image-level matching cannot identify which object in a cluttered image truly matches the composed expression. This issue becomes critical when multiple same-category objects appear together, when distractors match only part of the text, or when the desired output is an object mask rather than a whole-image retrieval result.

To facilitate object-level retrieval in complex scenes, we propose the \textbf{Composed Object Retrieval (\task{}) task}. As illustrated in \cref{fig:intro_motivation}(b), \task{} takes a target image with candidate objects, a reference image, a reference object mask, and a retrieval text as input. Given these components, \task{} retrieves and segments the most relevant object(s) from the candidate objects that match the composed expression formed by the reference object and retrieval text.
Specifically, each input component plays a distinct role. The reference mask specifies the reference object without relying on explicit class names, enabling flexible retrieval even when object categories are ambiguous or hard to describe. Meanwhile, the retrieval text describes the desired attribute modifications that distinguish the target object from the reference object. This enables \task{} to process complex expressions that require visual and textual reasoning.
Unlike traditional CIR that retrieves full images, \task{} retrieves object instances within a target scene and uses segmentation masks as pixel-level grounding of the retrieved results. This formulation preserves the match-and-select nature of retrieval while producing outputs useful for downstream object editing, inspection, and interaction.

\task{} is more challenging than CIR, as it requires retrieving precise objects that match complex composed expressions while carefully excluding similar but incorrect objects in the same scene.
Specifically, the task involves three main challenges:
\textbf{1) Compositional matching.} The model needs to understand the reference object and retrieval text jointly to capture subtle attribute changes such as color, shape, texture, or state.
\textbf{2) Negative object filtering.} The model needs to distinguish correct target objects from visually similar candidates in the same image that do not fully satisfy the composed expression.
\textbf{3) Multi-object retrieval.} The model must locate and segment one or more instances in the target image that match the composed expression.
These challenges are tightly coupled: a model must preserve reference-specific visual details, understand the linguistic modification, and compare the resulting composed representation against all candidate objects rather than independently segmenting salient regions.

To advance \task{} research, we construct \textbf{\dataset{}}, a large-scale benchmark built automatically using public images~\cite{lin2014microsoft,gupta2019lvis} and large multimodal models~\cite{Qwen2.5-VL}. Spanning 408 categories, it includes 125,541 retrieval triplets (\ie, target object, reference object, retrieval text) across 28,150 images and 35,584 objects, divided into Train, Test-Base, and Test-Novel subsets, with Test-Novel containing 78 novel categories to evaluate category-level generalization. The construction pipeline combines object filtering, reference-target pairing, retrieval-text generation, automatic verification, and human inspection, providing a reproducible benchmark for object-level composed retrieval.
We present an end-to-end baseline, \textbf{\model{}} (\textbf{C}omposed \textbf{O}bject \textbf{RE}trieval), which integrates three key components: \textbf{1) a Reference Region Embedding (RRE) module} that extracts region-level features from reference objects while preserving context; \textbf{2) an Adaptive Vision-Text Interaction (AVTI) module} that constructs composed representations through dynamic multimodal fusion; and \textbf{3) a \task{}-oriented contrastive loss} that aligns target features with composed expressions while suppressing background and distractors.
Experimental results show that \model{} achieves \sArt performance on \dataset{}, surpassing existing methods. It improves Dice by 34.9\% and IoU by 36\% on Test-Base, and by 20.7\% and 19\% on Test-Novel, respectively.
In summary, our main contributions are as follows.

\begin{itemize}
\item \textbf{New task.} We introduce \textbf{Composed Object Retrieval (\task{})}, a fine-grained task that extends composed image retrieval to instance-level retrieval with pixel-level grounding. It retrieves object instance(s) using a reference object and text modification, without category names.

\item \textbf{Large-scale benchmark.} We construct \textbf{\dataset{}}, the first large-scale \task{} benchmark, with 125,541 triplets across 408 categories, base/novel splits, and pixel-level masks for evaluation.

\item \textbf{Unified baseline.} We present \textbf{\model{}}, combining reference modeling, adaptive fusion, and contrastive learning for aligning composed representations with targets and suppressing negatives.

\item \textbf{Strong performance.} \model{} achieves \sArt results on \dataset{}, outperforming CIR-based pipelines and strong baselines in base/novel categories.

\end{itemize}

\section{Related Works}\label{sec:related_works}
\minisection{Composed Image Retrieval.}
Composed Image Retrieval (CIR) uses a reference image and text to retrieve targets by combining visual cues with textual attributes. Early methods used handcrafted or modality-specific fusion~\cite{vo2019composing,chen2020image,delmas2022artemis,dodds2020modality,kim2021dual}, while later ones adopted contrastive learning and attention for semantic alignment.
Current CIR pipelines~\cite{baldrati2023composed,liu2023candidate,xu2024sentence,feng2024improving,baldrati2022effective,gu2024language,yang2024semantic,sun2023training,kolouju2025good4cir,wang2025aligning,levy2024data,wen2023self,agnolucci2025isearle,zhou2025dual,li2026combiner,zhang2022composed,yang2023composed} typically use global encoders~\cite{baldrati2023composed,baldrati2022effective,radford2021learning} or fine-grained reasoning~\cite{liu2023candidate,xu2024sentence,feng2024improving}, followed by fusion through target-guided composition~\cite{baldrati2022effective,delmas2022artemis,wen2023target}, semantic editing~\cite{zhao2022progressive,yang2024semantic}, or bidirectional training~\cite{liu2023candidate,xu2024sentence}. Recent extensions include zero-shot adaptation~\cite{gu2024language}, re-ranking~\cite{sun2023training}, instruction-guided retrieval~\cite{zhang2024magiclens}, data enrichment~\cite{kolouju2025good4cir,levy2024data}, negative mining~\cite{wang2025aligning}, video retrieval~\cite{ventura2024covr}, diffusion augmentation~\cite{gu2023compodiff}, fine-grained parsing~\cite{li2025finecir}, action modeling~\cite{li2025encoder}, and consistency learning~\cite{xing2025context}.
Recent benchmarks advance CIR with synthetic pairs, LLM-assisted retrieval, multimodal arithmetic reasoning, and instance-level retrieval. However, most lack dense object masks or do not test whether composed expressions can identify and ground instances among distractors. Existing CIR therefore remains largely image-level and cannot precisely localize or distinguish objects in complex scenes. Moreover, datasets such as FashionIQ~\cite{Guo2021FashionIA} and CIRR~\cite{liu2021image} lack region or pixel-level annotations, preventing object-level evaluation. To address these limitations, we propose the \task{} task, extending multimodal retrieval to the object level.

\minisection{Pixel-Level Retrieval and Grounding.}
Beyond image-level retrieval, studies on pixel retrieval~\cite{an2023towards,kim2026pixel} and personalized retrieval~\cite{samuel2024s} with segmentation show that masks can serve as retrieval outputs or inputs, enabling retrieval to be grounded at the pixel or instance level. Our work follows this direction but focuses on composed object retrieval, where the composed expression consists of a reference object and a text modification, and the model must identify matching object instance(s) in a target image while suppressing distractors.

\minisection{Vision-Language Models.}
Vision–Language Models (VLMs) learn multimodal representations from large-scale data~\cite{radford2021learning,li2021align,li2022blip}. CLIP~\cite{radford2021learning} and ALBEF~\cite{li2021align} use contrastive alignment, while BLIP and BLIP-2~\cite{li2022blip,li2023blip} introduce encoder-decoder structures for stronger reasoning.
However, most VLMs operate at the image level, limiting object localization and segmentation. We therefore propose a unified framework that integrates SigLIP~\cite{zhai2023sigmoid} for vision–language alignment with SAM~\cite{kirillov2023segment} for object segmentation, bridging image-level semantics and pixel-level precision for the \task{}.

\begin{figure*}[t]
\centering
\includegraphics[width=\linewidth]{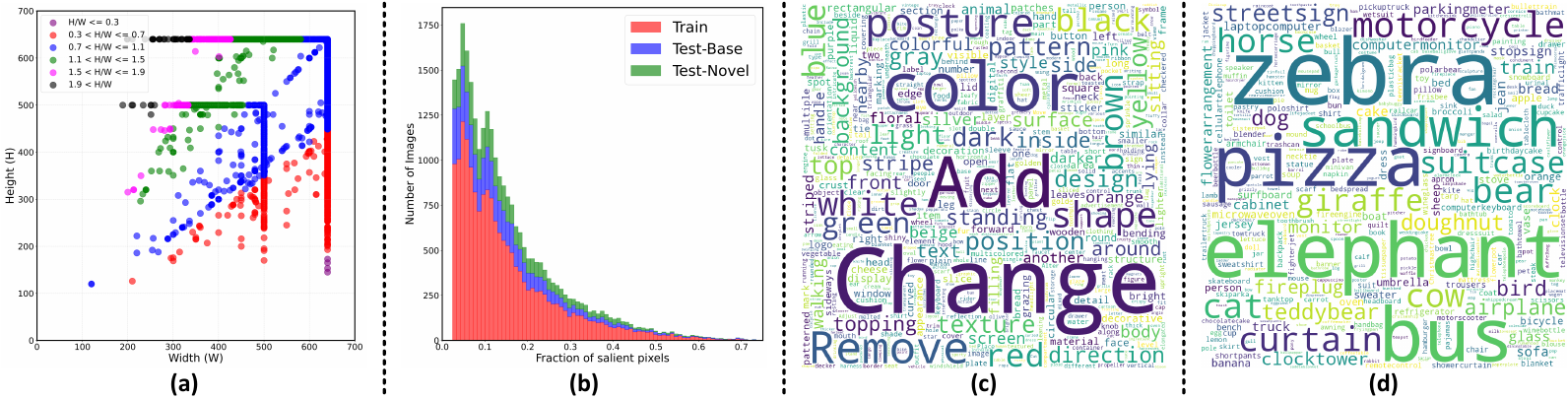}
\vspace{-8mm}
\caption{\textbf{\dataset{} is diverse in scale, text, and categories.}
(a) Distribution of image resolutions, with colors denoting resolution-ratio intervals to highlight scale diversity.
(b) Object-to-image area ratio, where colors indicate different subsets for direct comparison.
(c) Word cloud of retrieval texts and (d) category word cloud, with word size proportional to frequency, revealing common expressions and dominant categories.
}
\vspace{-4mm}
\label{fig:dataset_details}
\end{figure*}

\section{Composed Object Retrieval}\label{sec:composed_object_retrieval}
\subsection{Task Definition}
This paper introduces Composed Object Retrieval (\task{}), which retrieves one or more target objects $O_{tar}$ from candidate objects in a target image $I_{tar}$ using a composed expression formed by a reference object $O_{ref}$ in a reference image $I_{ref}$ and a retrieval text $T_{ret}$. Unlike conventional image-level retrieval over an external gallery, \task{} performs object-level retrieval within a target scene: it matches the composed expression to candidate objects, selects all satisfying instances, and grounds them with pixel-level masks.

The task takes four inputs: a reference image $I_{ref}$, a binary mask $M_{ref}$ specifying $O_{ref}$, a target image $I_{tar}$, and a retrieval text $T_{ret}$ describing the attribute-level transformation from $O_{ref}$ to $O_{tar}$. The output is a binary mask $M_{tar}$ that grounds the retrieved target object(s) in $I_{tar}$. For multi-object retrieval, $M_{tar}$ covers all target instances satisfying the composed expression. The retrieval text describes attribute-level changes (\eg, shape, color, spatial relations) without naming the object, enabling generalization to novel or ambiguous categories. In practice, $M_{ref}$ can be obtained through simple interactions such as points, boxes, or click-based segmentation, making \task{} practical for real-world use.

\task{} differs from existing tasks in three aspects.
\textbf{1) Compared with CIR}, which performs image-level matching, \task{} retrieves object instances and outputs segmentation masks for pixel-level grounding.
\textbf{2) Unlike text-only referring segmentation}~\cite{Ye2021ReferringSI,Liu2021CrossModalPC,Liu2023MultiModalMA,Wu2024TowardRR} , \task{} uses reference images and masks to visually specify the source object, while $T_{ret}$ describes the desired modification, reducing ambiguity and capturing fine-grained differences.
\textbf{3) Unlike standard segmentation}~\cite{Qi2024HighQualityES,Liu2017PiCANetLP,Wang2024PolypSV}, \task{} emphasizes candidate competition, requiring the model to distinguish correct target object(s) from same-category or semantically related distractors that partially match the composed expression.

\subsection{\dataset{} Dataset}\label{sec:data_details}
\minisection{\dataset{} Dataset Details.}
We present \dataset{}, a large-scale dataset for the \task{}, containing 125,541 retrieval triplets generated by a ten-step automated pipeline. Each triplet consists of a reference object $O_{ref}$, a retrieval text $T_{ret}$ describing attribute-level changes, and a target object $O_{tar}$.
Each object is localized by a binary mask: $M_{ref}$ defines $O_{ref}$ in $I_{ref}$, and $M_{tar}$ defines $O_{tar}$ in $I_{tar}$. For multi-object retrieval, $M_{tar}$ is the union mask of all target instances satisfying the composed expression, covering both single- and multi-instance outputs. Thus, each pair is represented as $(I_{ref}, M_{ref}, T_{ret}, I_{tar}, M_{tar})$, while retrieval triplet denotes the semantic relation $(O_{ref}, T_{ret}, O_{tar})$.
Overall, \dataset{} includes 28,150 images, 35,584 annotated objects, and 408 categories, forming a challenging benchmark for object-level retrieval.
We split \dataset{} into Train, Test-Base, and Test-Novel to evaluate seen-category performance and category-level generalization. Test-Base uses categories seen during training, while Test-Novel contains 78 held-out categories, separating instance-level generalization from transfer to unseen object types. The dataset supports single- and multi-object retrieval and includes visually similar distractors for fine-grained discrimination beyond coarse category matching.

\begin{table}[t]
\centering
\caption{Statistics of samples and categories in \dataset{}.}
\vspace{-2mm}
\label{tab:metrics_stats}
\large
\setlength{\tabcolsep}{6pt}
\renewcommand{\arraystretch}{0.9}
\resizebox{\linewidth}{!}{%
\begin{tabular}{lrrrr}
\toprule
Metric & All & Train & Test-Base & Test-Novel  \\
\midrule
Retrieval Triplets & 125,541 & 84,303 & 23,337 & 17,901  \\
Total Categories & 408 & 330 & 284 & 78  \\
Target Images $I_{tar}$ & 21,434 & 14,668 & 4,371 & 3,735  \\
Target Objects $O_{tar}$ & 26,576 & 17,497 & 4,921 & 4,158  \\
Reference Images $I_{ref}$ & 16,477 & 12,779 & 6,982 & 3,308  \\
Reference Objects $O_{ref}$ & 18,338 & 13,945 & 7,278 & 3,378  \\
All Images $I_{all}$ & 28,150 & 20,493 & 11,010 & 5,125  \\
All Objects $O_{all}$ & 35,584 & 24,685 & 11,949 & 5,637  \\
\bottomrule
\end{tabular}}
\vspace{-6mm}
\end{table}

\cref{tab:metrics_stats} summarizes the statistics of \dataset{} and its subsets: Train, Test-Base, and Test-Novel.
The Train set comprises 84,303 triplets with 17,497 target objects from 14,668 images and 13,945 reference objects from 12,779 images. Test-Base includes 23,337 triplets with 4,921 targets from 4,371 images and 7,278 references from 6,982 images, while Test-Novel contains 17,901 triplets with 4,158 targets from 3,735 images and 3,378 references from 3,308 images. Since one image may contain multiple annotated objects and one object may form multiple valid reference-text pairings, we report both image-level and object-level counts.
For each row, the ``All'' column reports globally unique entities after deduplication across splits; for $I_{all}$ and $O_{all}$, deduplication is further performed across target and reference roles.
\cref{fig:dataset_details} (a)–(d) further show target-to-image resolution ratios, object-to-image area ratios, and word clouds for retrieval texts and object categories. These statistics indicate that \dataset{} covers varied object scales and diverse textual transformations rather than a small set of templates.
We next summarize the data splits, followed by examples and category-level statistics.

\minisection{Data Splits and Pretraining Contamination.}
\dataset{} images are sourced from COCO 2017 train/val~\cite{lin2014microsoft} and LVIS v1~\cite{gupta2019lvis}, but all composed retrieval annotations are newly constructed, including reference masks, retrieval texts, target masks, and reference-target pairings.
To ensure fair evaluation, we enforce target-object-disjoint splits: no target object appears in more than one of Train, Test-Base, and Test-Novel, and all Test-Novel categories are unseen during training.
Although pretrained vision-language encoders such as CLIP, BLIP, and SigLIP may have encountered COCO images during pretraining, generic pretraining does not provide the composed objective of identifying the object that jointly matches a reference object and a retrieval text.
This is reflected in our experiments, where pretrained CIR baselines still require task-specific composed object alignment to perform well.

\minisection{Dataset Examples.}
\cref{fig:dataset_examples} illustrates nine representative retrieval triplets from \dataset{}.
Each example contains a reference image $I_{ref}$ and a target image $I_{tar}$ in the first row, followed by the highlighted reference object $O_{ref}$ and target object $O_{tar}$ in the second row.
The highlighted objects correspond to the reference and target masks used by COR, making the object-level retrieval relation explicit even when the original images contain multiple candidates or distractors.
The retrieval texts for examples \textbf{a}-\textbf{i} are: \textit{``Add green border''}, \textit{``Remove the food items''}, \textit{``Change the color to green''}, \textit{``Change the color to yellow''}, \textit{``Change the color to red''}, \textit{``Change the color to dark brown''}, \textit{``Add white patches on top''}, \textit{``Change the color to light blue''}, and \textit{``Add a handle on top''}, respectively.
These examples reflect the core challenges of \task{}: compositional matching, distractor suppression, and multi-object retrieval.
\begin{figure}[t]
    \centering
    \includegraphics[width=\linewidth]{figures/appendix_fig_3_dataset_example_compressed.pdf}
    \vspace{-8mm}
    \caption{\textbf{\dataset{} provides object-level composed retrieval triplets.} We present nine retrieval triplets. For each example, the first row shows the original reference and target images, and the second row highlights the corresponding reference and target objects.}
    \vspace{-6mm}
    \label{fig:dataset_examples}
\end{figure}

\begin{figure}[!htb]
\vspace{-4mm}
\centering
\includegraphics[width=\linewidth]{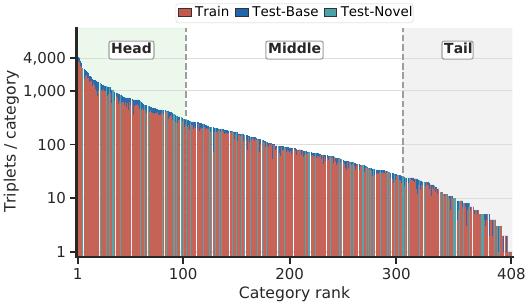}
\vspace{-8mm}
\caption{\textbf{\dataset{} has a long-tailed rank-frequency distribution.} Categories are sorted by the total number of retrieval triplets. Shaded regions indicate head, middle, and tail category groups; colors denote Train, Test-Base, and Test-Novel subsets.
}
\vspace{-4mm}
\label{fig:dataset_histogram}
\end{figure}

\begin{figure}[!htb]
\vspace{-2mm}
\centering
\includegraphics[width=\linewidth]{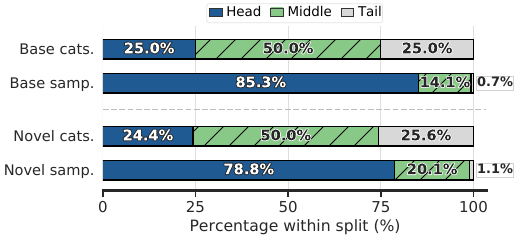}
\vspace{-8mm}
\caption{\textbf{Head categories dominate the sample mass in evaluation splits.} Category and sample mass are shown under the head/middle/tail grouping for Test-Base and Test-Novel. Percentages are normalized within each split.
}
\vspace{-4mm}
\label{fig:dataset_histogram_2}
\end{figure}

\minisection{Category Distribution.}
Beyond individual examples, \dataset{} covers 408 object categories and exhibits a pronounced long-tailed frequency distribution, as shown in \cref{fig:dataset_histogram}.
Categories are ranked by the total number of retrieval triplets rather than listed alphabetically, making the head-to-tail decay explicit.
The shaded regions define the head (top 25\%), middle (middle 50\%), and tail (bottom 25\%) groups, while the colored curves show how Train, Test-Base, and Test-Novel samples are distributed across the same ranked category axis.
\cref{fig:dataset_histogram_2} further summarizes the category mass and sample mass of the head/middle/tail grouping within each evaluation split, showing that a small number of head categories account for most retrieval triplets whereas many tail categories contribute relatively few samples.
This design supports evaluation across frequent, middle-frequency, rare, and unseen categories.

\minisection{Automated Annotation Generation.}
To construct \dataset{}, we use an automated pipeline based on COCO 2017~\cite{lin2014microsoft}, LVIS~\cite{gupta2019lvis}, and Qwen2.5-VL~\cite{Qwen2.5-VL}.
The pipeline integrates image filtering, data splitting, triplet construction, retrieval-text generation, and quality control into \textbf{four stages with ten steps}, producing 125,541 semantically consistent triplets.
Stage 1 establishes a clean object pool, Stage 2 creates base/novel and train/test splits, Stage 3 builds reference-target-text triplets, and Stage 4 removes ambiguous or inconsistent samples through retrieval verification and false match rejection.

\begin{figure}[!htbp]
\vspace{-2mm}
  \centering
  \includegraphics[width=\linewidth]{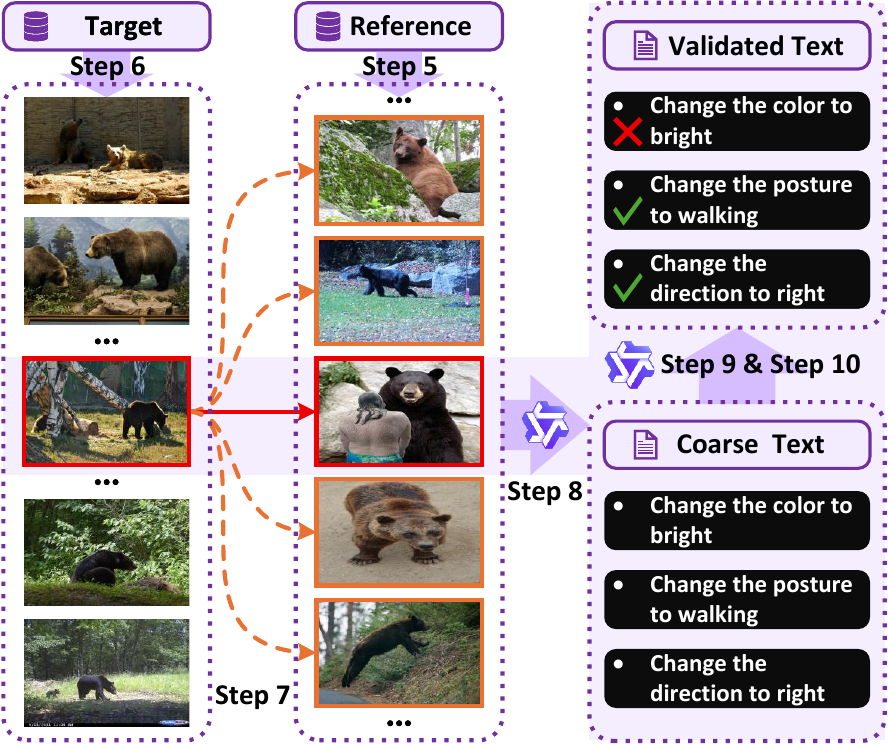}
  \vspace{-6mm}
  \caption{\textbf{Pair sampling and validation construct reliable retrieval triplets.}
  The figure illustrates reference selection, target sampling, retrieval-text generation, and ambiguity filtering.
  \vspace{-2mm}
  }\label{fig:dataset_annotation}
\end{figure}

\minisection{Detailed Annotation Pipeline.}\label{sec:dataset_annotation}
\cref{fig:dataset_annotation} illustrates the core triplet-construction and validation process after raw data preprocessing and data splitting.
In the target image, same-category objects can act as hard distractors; for example, the shown scene contains a brown bear as the positive target and a black bear as a negative object.
The pipeline first selects a clear same-category reference object that is visually distinguishable from the target, then constructs a reference--target pair suitable for generating a composed expression.
Given the verified pair, the LLM generates a retrieval text describing the transformation from the reference object to the target object.
However, vague descriptions such as \textit{``Change the color to bright''} may not accurately capture the intended change or may still match a distractor.
Therefore, retrieval verification and false match rejection remove ambiguous, inconsistent, or non-unique triplets.
Below, we summarize the complete annotation pipeline and retain the prompts used for automatic quality control, retrieval-text generation, and triplet validation.

\minisection{Stage 1: Raw Data Preprocessing.}
\textit{Step 1} filters COCO 2017 images. We remove instances occupying less than 3\% or more than 80\% of the image area, keep masks covering at least 20\% of the bounding box, discard categories with fewer than two images or more than three same-category objects per image, and cap categories with more than 300 valid samples. \textit{Step 2} uses Qwen2.5-VL to remove objects with severe occlusion, blur, truncation, or poor recognizability.

\vspace{-1mm}
\begin{tcolorbox}[
    myboxstyle,
    title=Stage 1 Prompts: Raw Data Preprocessing
]
\textbf{Step 2: Low-Quality Object Removal.}

You are a professional visual data quality control expert.
Please carefully evaluate whether the given \texttt{[IMAGE]} contains valid and high-quality instances of the target category \texttt{\{cat\_name\}} according to the following rules:

(1) There must be at least \texttt{\{ins\_len\}} clearly visible instances of \texttt{\{cat\_name\}} in the image.  \\
(2) Each \texttt{\{cat\_name\}} instance should appear complete and not truncated by the image border.  \\
(3) The instances must not be blurry, heavily occluded, or indistinguishable from the background.  \\
(4) The objects should maintain reasonable visual scale and clarity to allow recognition by a human annotator.  \\
If all conditions are satisfied, return \textbf{1}; otherwise, return \textbf{0}.
Do not provide any additional explanation or text beyond the single digit output.
\end{tcolorbox}
\vspace{-1mm}

\minisection{Stage 2: Data Split.}
\textit{Step 3} divides categories into 330 base classes and 78 novel classes using a 4:1 ratio. \textit{Step 4} assigns all novel-category samples to Test-Novel, initially splits base-category samples into Train and Test-Base with a 3:1 ratio, and then removes training rows whose target objects overlap with evaluation targets. This produces target-object-disjoint training and testing subsets and avoids category leakage for Test-Novel.

\minisection{Stage 3: Retrieval Triplet Building.}
\textit{Step 5} selects reference objects that appear as a single instance and occupy at least 5\% of the image area. \textit{Step 6} groups target images into 1p0n, 1p1n, 1p2n, 2p0n, 2p1n, and 3p0n configurations. For negative-object settings, DINOv2~\cite{oquab2023dinov2} removes positive-negative pairs with cosine similarity above 0.8, and Qwen2.5-VL further checks whether positives and negatives are visually distinguishable in attributes such as color, shape, pose, or action. \textit{Step 7} pairs each target with up to five references and keeps only pairs whose reference and target are complete, clear, and distinguishable. \textit{Step 8} generates concise retrieval texts in the format \texttt{[(change1), (change2), (change3)]}. The generated text avoids category names and describes dynamic attributes for animals or static attributes for inanimate objects.

\vspace{-1mm}
\begin{tcolorbox}[
    myboxstyle,
    title=Stage 3 Prompts: Retrieval Triplet Building
]
\textbf{Step 6: Target Object Selection.}

You are a data quality control expert. Please evaluate whether the input \texttt{[IMAGE]} meets the following conditions:
(1) The \texttt{\{cat\_name\}} object marked with a red bounding box (positive sample) is visually distinguishable from the \texttt{\{cat\_name\}} object marked with a blue bounding box (negative sample).\\
(2) The differences should relate to attributes such as color, shape, pose, action, or overall appearance, while both objects must remain clearly visible and complete.\\
(3) The red-box object should not be occluded, truncated, or blurry.\\
Return \textbf{1} if all conditions are met; otherwise, return \textbf{0}. Do not include any additional explanation or text.

\medskip
\textbf{Step 7: Pair Construction and Verification.}

You are a data quality control expert. Two images are provided.\\
The \texttt{[IMAGE]} (reference) contains a \texttt{\{cat\_name\}} object marked with a red bounding box.\\
The \texttt{[IMAGE]} (target) contains a \texttt{\{cat\_name\}} object also marked with a red bounding box.\\
Please verify the following: \\
(1) The reference and target objects are distinguishable in attributes such as color, shape, pose, or action.\\
(2) Both objects are complete, clear, and not blurred or occluded.\\
(3) The target object can be uniquely identified even if other same-category objects exist in the image.\\
Return \textbf{1} if all the above conditions are met; otherwise, return \textbf{0}.

\medskip
\textbf{Step 8: Retrieval Text Generation.}

You are a visual data annotation expert. Two images are provided: \texttt{[IMAGE]} (reference) and \texttt{[IMAGE]} (target).
Each image contains a \texttt{\{cat\_name\}} object marked with a red bounding box, and the target image may also include other same-category objects as distractors.\\
Describe the attribute changes from \texttt{[IMAGE]} (reference) to \texttt{[IMAGE]} (target) so that the target object can be uniquely identified.\\
(1) For animals: focus on dynamic attributes such as pose, action, appearance, color, pattern, or direction.\\
(2) For inanimate objects: focus on static attributes such as shape, position, color, or layout.\\
Avoid using category names or the words ``reference'' and ``target''. Use clear, simple, and natural language.\\
Output the result strictly in the format \texttt{[(change1), (change2), (change3)]}, with each phrase no longer than ten words.
\end{tcolorbox}
\vspace{-1mm}

\minisection{Stage 4: Triplets Validation.}
\textit{Step 9} verifies that the reference object and retrieval text correctly identify the target object. \textit{Step 10} masks out the target object and asks Qwen2.5-VL whether any remaining object can still satisfy the same composed expression. Triplets are retained only when the positive match is valid and no false match is detected.

\vspace{-1mm}
\begin{tcolorbox}[
    myboxstyle,
    title=Stage 4 Prompts: Triplets Validation
]
\textbf{Step 9: Retrieval Verification.}

You are a data quality control expert. Two images are provided: \texttt{[IMAGE]} (reference) and \texttt{[IMAGE]} (target), each containing a \texttt{\{cat\_name\}} object marked with a red bounding box. The attribute change description is given as \texttt{\{retrieval\_text\}}.\\
Determine whether the target object in the second image matches the attribute changes described from the reference object in the first image.
Return \textbf{1} if the description accurately matches the target; otherwise, return \textbf{0}.

\medskip
\textbf{Step 10: False Match Rejection.}

You are a data quality control expert. Two images are provided: \texttt{[IMAGE]} (reference) and \texttt{[IMAGE]} (target).
In the second image, the target object has been masked out. The attribute change description is given as \texttt{\{retrieval\_text\}}.\\
Verify whether any remaining object in the masked image could still match the description when combined with the reference object.
Return \textbf{1} if no false matches are found and the retrieval text uniquely identifies the target object; otherwise, return \textbf{0}.
\end{tcolorbox}
\vspace{-1mm}

\minisection{Human Validation Protocol.}
To improve \dataset{} reliability beyond automatic validation, we conduct human inspection at multiple stages.
During pipeline development, we manually verify about 600 samples to refine filtering rules, retrieval-text generation, and quality-control prompts before scaling.
After construction, we randomly inspect about 600 final-test triplets for object visibility, mask quality, text-target consistency, and uniqueness under distractors.
Auditing challenging samples shows that most failures stem from subtle attribute differences, visually similar distractors, or multi-object ambiguity rather than systematic annotation noise.

\section{Approach}
\subsection{Limitations of Existing CIR Methods}
Current CIR methods are inadequate for \task{} due to: 1) they retrieve whole images rather than localizing specific objects; 2) they use global image features rather than the specified reference object, leading to suboptimal results.
Although a multistage pipeline combining CIR with detection and segmentation could handle \task{}, it has three drawbacks: 1) it is not end-to-end trainable, 2) it is computationally costly, and 3) it lacks multi-object retrieval support.

\subsection{Model Architecture}
\begin{figure*}[t]
\centering
\includegraphics[width=0.95\linewidth]{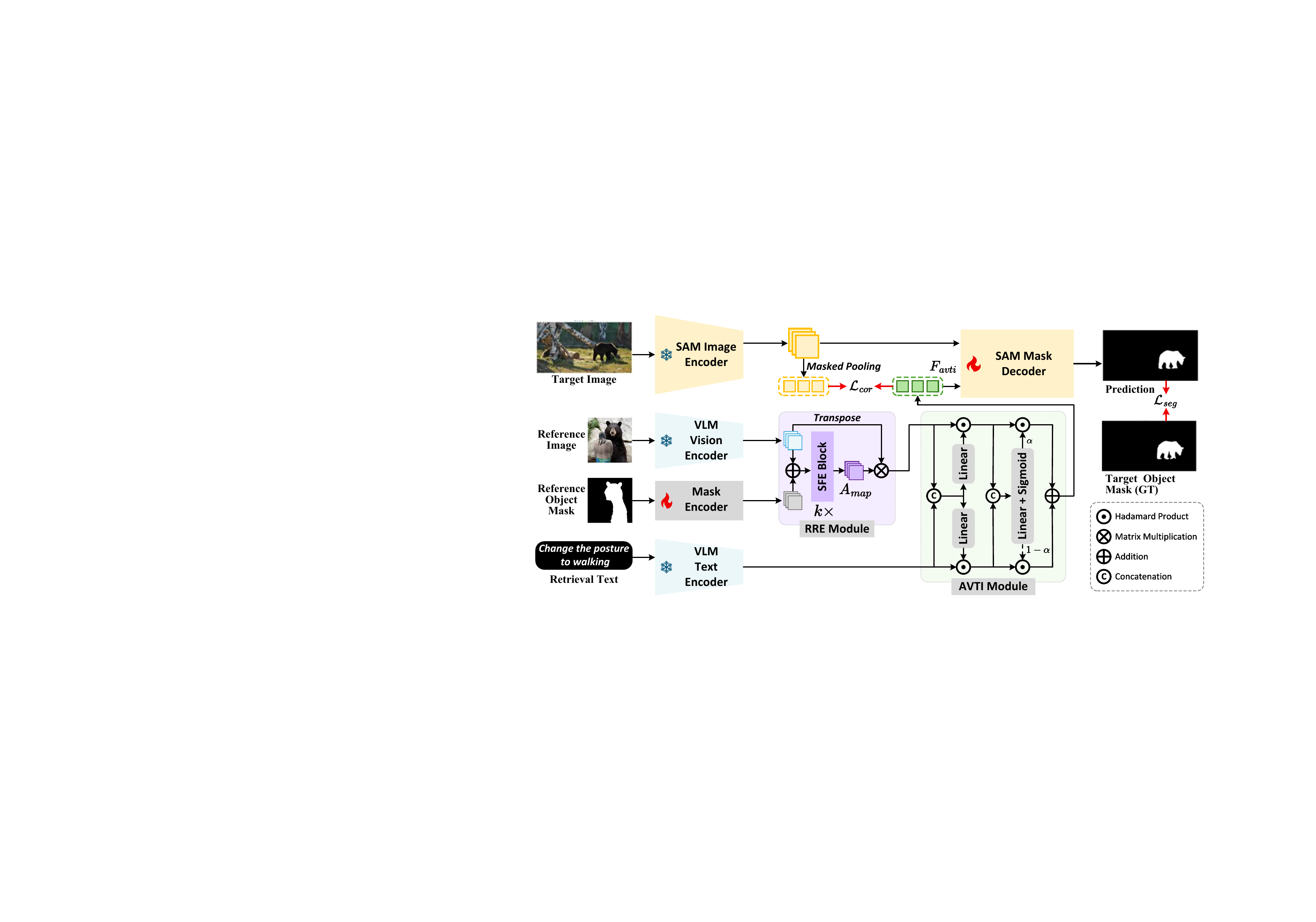}
\vspace{-3mm}
\caption{\textbf{\model{} performs composed object retrieval in an end-to-end framework.} Given a reference image, reference mask, retrieval text, and target image, the architecture uses the Reference Region Embedding (RRE) module to extract reference features and the Adaptive Vision-Text Interaction (AVTI) module to form a composed representation. A \task{}-oriented contrastive loss $\mathcal{L}_{cor}$ aligns target features with the composed expression while suppressing distractors.}
\vspace{-4mm}
\label{fig:model_arch}
\end{figure*}

\minisection{Overall.}
We propose \model{} (Composed Object REtrieval), an end-to-end baseline for \task{}, as illustrated in \cref{fig:model_arch}. \model{} integrates three core designs:
1) a Reference Region Embedding (RRE) module that extracts object-level features from reference images;
2) an Adaptive Vision-Text Interaction (AVTI) module that fuses reference visual features with retrieval text to produce discriminative composed representations; and 3) a \task{}-oriented contrastive loss $\mathcal{L}_{cor}$ that aligns target features with the composed expression while suppressing background and distractors.

The framework processes target image $I_{tar}$, reference image $I_{ref}$, reference object mask $M_{ref}$, and retrieval text $T_{ret}$ through the SAM image encoder, VLM vision encoder, mask encoder, and VLM text encoder, generating features $F_{tar}$, $F_{ref}$, $F_{mask}$, and $F_{txt}$. The RRE module combines a projected reference feature with $F_{mask}$ to estimate semantic activation maps and aggregate $F_{ref}$ into the reference representation $F_{rre}$. The AVTI module then fuses $F_{rre}$ with $F_{txt}$ to generate the composed representation $F_{avti}$, which is projected to the SAM sparse prompt token $\mathbf{f}_{avti}$ and guides the mask decoder to process $F_{tar}$ for the final prediction.

\minisection{Reference Region Embedding (RRE).}\label{sec:rre_module}
We utilize a mask to highlight the object of interest.
This strategy avoids class names, making it effective when categories are hard to define or require expert knowledge.
Instead of commonly used mask cropping~\cite{ding2022decoupling,liang2023open} or mask pooling~\cite{ghiasi2022scaling}, RRE learns to fuse mask features with image features, preserving semantic context and avoiding disruptions to the image distribution.
Inspired by~\cite{li2025mask}, RRE employs a semantic activation-based strategy that computes activation maps from reference image features and object masks, highlighting identity-relevant regions while preserving context.
Concretely, a mask encoder maps the reference mask $M_{ref}$ to the mask feature $F_{mask}$:
\begin{equation}\label{eq:RRE_EQ1}
\vspace{-1mm}
F_{mask} = \text{MaskEncoder}(M_{ref}). 
\vspace{-1mm}
\end{equation}

The RRE module first projects the reference image feature $F_{ref}$ into a channel-compatible feature $F'_{ref}$ and then fuses it with $F_{mask}$. The fused feature is processed by three stacked Semantic Feature Enhancing (SFE) blocks to enrich the semantic representation:
\begin{equation}\label{eq:RRE_EQ2}
\vspace{-1mm}
\text{SFE}(x) = x + \text{PWC}(\text{GELU}(\text{PWC}(\text{LN}(\text{DWC}(x))))),\\
\vspace{-1mm}
\end{equation}
where $x = F_{mask} + F'_{ref}$, $\text{DWC}$ is a $7 \times 7$ depth-wise convolution, $\text{PWC}$ is a $1 \times 1$ point-wise convolution, and $\text{LN}$ is layer normalization.
The semantic activation map $A_{map} \in \mathbb{R}^{h \times w \times K}$ is computed as:
\begin{equation}\label{eq:RRE_EQ3}
\vspace{-1mm}
A_{map} = \text{Conv}_{1\times 1}(\text{SFE}_3(x)),
\vspace{-1mm}
\end{equation}
where $h$ and $w$ denote the spatial resolution of the activation map, and $K$ denotes the number of semantic subspaces.

We obtain $F_{rre}$ by aggregating semantic-aware features via batch matrix multiplication between spatially flattened $F_{ref}$ and normalized activation maps $\bar{A}_{map}$:
\begin{equation}\label{eq:RRE_EQ4}
\vspace{-1mm}
F_{rre} = \frac{1}{K} \sum_{k=1}^K \bar{A}_{map}^{k} \cdot F_{ref}^T,
\vspace{-1mm}
\end{equation}
where $\bar{A}_{map}^{k}$ is the $k$-th normalized activation map obtained by applying spatial softmax to $A_{map}^{k}$. Averaging across $K$ subspaces yields $F_{rre}$, capturing various semantic cues for robust reference object representation.

\minisection{Adaptive Vision-Text Interaction (AVTI).}\label{sec:avti_module}
The AVTI module, inspired by \cite{baldrati2023composed,huang2024dynamic}, enhances the \task{} task by adaptively fusing reference object features with retrieval text to produce semantically rich embeddings. Given reference object features $F_{rre} \in \mathbb{R}^d$ and retrieval text features $F_{txt} \in \mathbb{R}^d$, where $d$ denotes the feature dimension, the module concatenates them into $F_{comb} = [F_{rre}, F_{txt}] \in \mathbb{R}^{2d}$. Modality-specific attention weights are computed via:
\begin{align}
\vspace{-0.5mm}
\text{attn}_V &= \sigma(\text{Linear}_{v2}(\text{ReLU}(\text{Linear}_{v1}(F_{comb})))), \label{eq:AVTI_EQ1} \\
\text{attn}_T &= \sigma(\text{Linear}_{t2}(\text{ReLU}(\text{Linear}_{t1}(F_{comb})))), \label{eq:AVTI_EQ2}
\vspace{-0.5mm}
\end{align}
where $\sigma(\cdot)$ is the sigmoid function. The attended features $\text{attn}_V \cdot F_{rre}$ and $\text{attn}_T \cdot F_{txt}$ are concatenated and processed for a scalar weight $\alpha \in [0, 1]$:
\begin{equation}\label{eq:AVTI_EQ3}
\vspace{-0.5mm}
\alpha = \sigma(\text{Linear}(\text{ReLU}(\text{Linear}([\text{attn}_V \cdot F_{rre}, \text{attn}_T \cdot F_{txt}])))).
\vspace{-0.5mm}
\end{equation}
The composed feature $F_{avti}$ is then generated as:
\begin{equation}\label{eq:AVTI_EQ4}
\vspace{-1mm}
F_{avti} = \alpha \cdot \text{attn}_V \cdot F_{rre} + (1 - \alpha) \cdot \text{attn}_T \cdot F_{txt},
\vspace{-1mm}
\end{equation}
We project $F_{avti}$ to a SAM-compatible sparse prompt token $\mathbf{f}_{avti}$, which guides the SAM mask decoder to process $F_{tar}$. The decoder uses $\mathbf{f}_{avti}$ to focus on the target object and produces prediction logits $F_{pred} = \text{MaskDecoder}(F_{tar}, \mathbf{f}_{avti})$, which are converted into the final target mask while suppressing background and distractor interference.

\minisection{Loss Function.}\label{sec:loss_func}
To enhance target distinction from background clutter, we propose a COR-oriented contrastive loss $\mathcal{L}_{cor}$, combining foreground alignment and background repulsion to align target features with the composed prompt and suppress non-target areas.
The foreground term ensures semantic consistency:
\begin{equation}\label{eq:LOSS_FG}
\vspace{-1mm}
\mathcal{L}_{fg} = 1 - \mathrm{CosSim}(F_{tar}^{fg}, \mathbf{f}_{avti}),
\vspace{-1mm}
\end{equation}
where $M_{gt}$ denotes the ground-truth target mask, $F_{tar}^{fg} = \mathrm{MaskedPooling}(F_{tar}, M_{gt})$ is the masked average-pooled foreground feature, and $\mathrm{CosSim}(a, b) = \frac{a \cdot b}{\|a\| \|b\|}$ denotes cosine similarity.
The background repulsion term minimizes similarity with distractors:
\begin{equation}\label{eq:LOSS_BG}
\vspace{-1mm}
\mathcal{L}_{bg} = 1 + \mathrm{CosSim}(F_{tar}^{bg}, \mathbf{f}_{avti}),
\vspace{-1mm}
\end{equation}
where $F_{tar}^{bg} = \mathrm{MaskedPooling}(F_{tar}, 1 - M_{gt})$ represents the background feature, reducing distractor influence.
The COR-oriented contrastive loss is $\mathcal{L}_{cor} = \mathcal{L}_{fg} + \mathcal{L}_{bg}$.

The final training objective integrates the contrastive and segmentation losses as
$\mathcal{L}_{total} = \mathcal{L}_{seg} + \mathcal{L}_{cor}$,
where $\mathcal{L}_{seg} = \mathcal{L}_{wbce} + \mathcal{L}_{wiou}$. Here, $\mathcal{L}_{wbce}$ and $\mathcal{L}_{wiou}$ denote edge-aware weighted binary cross-entropy and IoU losses, respectively.

\subsection{Implementation Details}\label{subsec:implementation_details}
\minisection{Training Protocol.}
We instantiate the target-image encoder and mask decoder with SAM-Base~\cite{kirillov2023segment}, and use SigLIP-Base~\cite{zhai2023sigmoid} for reference-image and text encoding.
All pre-trained backbones are frozen; only the new mask encoder, reference feature projection, RRE, AVTI, and task-specific prediction components are optimized.
We fine-tune the model for 15 epochs using AdamW with a learning rate of $1\times10^{-4}$.
The input resolution is $1024\times1024$ for SAM and $384\times384$ for SigLIP.
Training is performed on 4 RTX 4090 GPUs with BF16 precision and a per-GPU batch size of 6.

\minisection{RRE Implementation.}
The SigLIP vision encoder encodes the reference image, producing $F_{ref}\in\mathbb{R}^{768\times24\times24}$.
For mask-branch compatibility, a $1\times1$ convolution followed by Layer Normalization and GELU projects $F_{ref}$ to $F'_{ref}\in\mathbb{R}^{256\times24\times24}$.
The reference mask $M_{ref}\in\mathbb{R}^{1\times384\times384}$ is resized to the same spatial grid and passed through a lightweight convolutional mask encoder, whose final $1\times1$ convolution produces $F_{mask}\in\mathbb{R}^{256\times24\times24}$.
We fuse $F'_{ref}$ and $F_{mask}$ by element-wise addition and refine the result with three SFE blocks.
The refined feature generates $K=16$ semantic activation maps, each normalized by $\log\sigma(\cdot)$ followed by a spatial softmax.
These activation maps aggregate the original 768-channel visual feature $F_{ref}$, and the averaged subspace representation forms $F_{rre}\in\mathbb{R}^{1\times768}$.
Thus, RRE uses the mask to estimate object-aware spatial weights while preserving the richer semantic channels from the VLM vision encoder.

\minisection{SAM Decoder Prompt.}
The retrieval text is encoded by the SigLIP text encoder into $F_{txt}\in\mathbb{R}^{768}$.
AVTI fuses $F_{rre}$ and $F_{txt}$ to obtain $F_{avti}$ and projects it into a single SAM-compatible sparse prompt token $\mathbf{f}_{avti}\in\mathbb{R}^{1\times256}$.
During decoding, $\mathbf{f}_{avti}$ is provided to the SAM mask decoder as a sparse prompt embedding.
We use lowercase bold symbols for a single token vector and uppercase bold symbols for token sequences.
Following the original SAM decoder design, it is appended to the learnable output-token sequence composed of an IoU token $\mathbf{z}_{\mathrm{iou}}\in\mathbb{R}^{1\times256}$ and $N_m$ mask tokens $\mathbf{Z}_{\mathrm{mask}}\in\mathbb{R}^{N_m\times256}$:
\begin{equation}
\vspace{-1mm}
\mathbf{Z}_{\mathrm{dec}} = \operatorname{Concat}_{\mathrm{tok}}(\mathbf{z}_{\mathrm{iou}}, \mathbf{Z}_{\mathrm{mask}}, \mathbf{f}_{avti})
\in\mathbb{R}^{(N_m+2)\times256},
\vspace{-1mm}
\end{equation}
where $\operatorname{Concat}_{\mathrm{tok}}$ denotes concatenation along the token dimension.
The resulting decoder-token sequence attends to the target image feature $F_{tar}$ through the Transformer decoder.
No additional point, box, or dense prompt is used, and the remaining SAM components are kept unchanged.
The decoder outputs $F_{pred}\in\mathbb{R}^{1\times256\times256}$, which is converted into the final target object mask.

\subsection{Model Algorithm}
Algorithm~\ref{alg:core} summarizes \model{} inference. Given a reference image, reference mask, retrieval text, and target image, the model extracts modality-specific features, uses RRE to encode the reference object, applies AVTI to build a SAM-compatible composed prompt, and decodes the target mask with the SAM mask decoder. This keeps the composed expression and target-scene grounding in an end-to-end pipeline.

\begin{algorithm2e}[t]
\renewcommand{\arraystretch}{0.95}
\SetAlgoLined
\SetKwInOut{KwInput}{Input}
\SetKwInOut{KwOutput}{Output}
\KwInput{
Reference image $I_{ref}$, reference mask $M_{ref}$, target image $I_{tar}$, retrieval text $T_{ret}$.\\
VLM vision encoder $\mathcal{E}_{vis}$, VLM text encoder $\mathcal{E}_{txt}$, mask encoder $\mathcal{E}_{mask}$, SAM image encoder $\mathcal{E}_{sam}$, SAM mask decoder $\mathcal{D}_{sam}$.}
\KwOutput{Target object mask $M_{tar}$.}

\textbf{Feature Extraction:}\\
$F_{ref} = \mathcal{E}_{vis}(I_{ref}) \in \mathbb{R}^{768\times24\times24}$;\\
$F_{mask} = \mathcal{E}_{mask}(M_{ref}) \in \mathbb{R}^{256\times24\times24}$;\\
$F_{txt} = \mathcal{E}_{txt}(T_{ret}) \in \mathbb{R}^{768}$;\\
$F_{tar} = \mathcal{E}_{sam}(I_{tar}) \in \mathbb{R}^{256\times64\times64}$;\\[2pt]

\textbf{Reference Region Embedding (RRE):}\\
$F'_{ref} = \text{Reduce}(F_{ref}) \in \mathbb{R}^{256\times24\times24}$;\\
$x = F'_{ref} + F_{mask}$;\\
$A_{map} = \text{Conv}_{1\times1}(\text{SFE}_3(x))$;\\
$F_{rre} = \frac{1}{K} \sum_{k=1}^{K} \bar{A}_{map}^{k}\cdot F_{ref}^{T}$;\\[2pt]

\textbf{Adaptive Vision-Text Interaction (AVTI):}\\
$F_{comb} = [F_{rre}, F_{txt}]$;\\
$\text{attn}_V = \sigma(\text{Linear}_{v2}(\text{ReLU}(\text{Linear}_{v1}(F_{comb}))))$;\\
$\text{attn}_T = \sigma(\text{Linear}_{t2}(\text{ReLU}(\text{Linear}_{t1}(F_{comb}))))$;\\
$\alpha = \sigma(\text{Linear}(\text{ReLU}(\text{Linear}([\text{attn}_V\!\cdot\!F_{rre},\,\text{attn}_T\!\cdot\!F_{txt}]))))$;\\
$F_{avti} = \alpha\,\text{attn}_V\!\cdot\!F_{rre} + (1-\alpha)\,\text{attn}_T\!\cdot\!F_{txt}$;\\
$\mathbf{f}_{avti} = \text{Proj}(F_{avti}) \in \mathbb{R}^{1\times256}$;\\[2pt]

\textbf{Segmentation Decoding (SAM Head):}\\
$\mathbf{z}_{\mathrm{iou}},\mathbf{Z}_{\mathrm{mask}} \leftarrow \text{learnable SAM output tokens}$;\\
$\mathbf{Z}_{\mathrm{dec}} = \operatorname{Concat}_{\mathrm{tok}}(\mathbf{z}_{\mathrm{iou}},\mathbf{Z}_{\mathrm{mask}},\mathbf{f}_{avti})$;\\
$F_{pred} = \mathcal{D}_{sam}(F_{tar}, \mathbf{Z}_{\mathrm{dec}})$;\\[2pt]

\textbf{Output:} $M_{tar} = \text{Sigmoid}(F_{pred})$, where $F_{pred} \in \mathbb{R}^{1\times256\times256}$.
\caption{\textbf{CORE:} The model integrates RRE for object-aware reference encoding, AVTI for adaptive vision-text fusion, and SAM decoding guided by the composed prompt $\mathbf{f}_{avti}$ to predict the target object.}
\label{alg:core}

\end{algorithm2e}

\section{Experiments}\label{sec:experiments}
\begin{table*}[!t]
\centering
\caption{Quantitative results. \textbf{Bold} indicates the best, \underline{underline} denotes the second-best, and percentages show gains over the second-best. Higher Dice, IoU, mDice, and mIoU indicate better performance, while lower MAE denotes more accurate results.}
\vspace{-2mm}
\label{tab:baseline}
\setlength{\tabcolsep}{2pt}
\renewcommand{\arraystretch}{1.1}
\resizebox{\linewidth}{!}{
\begin{tabular}{l | c | ccccc | ccccc}
\toprule
\multirow{2}{*}{Method} & \multirow{2}{*}{Year} & \multicolumn{5}{c|}{\dataset{}-Test-Base} & \multicolumn{5}{c}{\dataset{}-Test-Novel} \\
 &  & Dice $\uparrow$ & IoU $\uparrow$ & MAE $\downarrow$ & mDice $\uparrow$ & mIoU $\uparrow$
 & Dice $\uparrow$ & IoU $\uparrow$ & MAE $\downarrow$ & mDice $\uparrow$ & mIoU $\uparrow$ \\
\midrule
CLIP4CIR~\cite{baldrati2023composed} & 2023 & 0.5333 & 0.4771 & 0.1166 & 0.7292 & 0.6759 & 0.5420 & 0.4903 & 0.1149 & 0.7347 & 0.6842 \\
BLIP4CIR~\cite{liu2023candidate} & 2023 & 0.5146 & 0.4570 & 0.1251 & 0.7174 & 0.6618 & 0.5032 & 0.4462 & 0.1306 & 0.7107 & 0.6545 \\
BLIP24CIR~\cite{xu2024sentence} & 2024 & 0.5157 & 0.4585 & 0.1180 & 0.7203 & 0.6660 & 0.5097 & 0.4546 & 0.1179 & 0.7184 & 0.6649 \\
Bi-BLIP4CIR~\cite{liu2024bi} & 2024 & 0.5308 & 0.4729 & 0.1231 & 0.7258 & 0.6706 & 0.5490 & 0.4916 & 0.1207 & 0.7364 & 0.6818 \\
CLIP4CIR-SPN~\cite{feng2024improving} & 2024 & 0.5474 & 0.4895 & 0.1137 & 0.7371 & 0.6835 & 0.5545 & 0.5004 & 0.1120 & 0.7419 & 0.6905 \\
BLIP4CIR-SPN~\cite{feng2024improving} & 2024 & 0.5277 & 0.4692 & 0.1232 & 0.7243 & 0.6687 & 0.5222 & 0.4644 & 0.1270 & 0.7211 & 0.6652 \\
BLIP24CIR-SPN~\cite{feng2024improving} & 2024 & 0.5709 & 0.5094 & 0.1098 & 0.7501 & 0.6952 & \underline{0.5883} & \underline{0.5285} & \underline{0.1044} & \underline{0.7612} & \underline{0.7081} \\
CompoDiff~\cite{gu2023compodiff} & 2024 & 0.5446 & 0.4871 & 0.1141 & 0.7355 & 0.6821 & 0.5501 & 0.4923 & 0.1125 & 0.7403 & 0.6863 \\
ENCODER~\cite{li2025encoder} & 2025 & 0.5530 & 0.4943 & 0.1128 & 0.7402 & 0.6863 & 0.5592 & 0.5001 & 0.1103 & 0.7455 & 0.6917 \\
ConText-CIR~\cite{xing2025context} & 2025 & 0.5615 & 0.5014 & 0.1114 & 0.7449 & 0.6905 & 0.5702 & 0.5102 & 0.1080 & 0.7528 & 0.6985 \\
FineCIR~\cite{li2025finecir} & 2025 & \underline{0.5712} & \underline{0.5114} & \underline{0.1088} & \underline{0.7521} & \underline{0.6972} & 0.5835 & 0.5221 & 0.1052 & 0.7600 & 0.7054 \\
\rowcolor{c1!30}
\textbf{\model{} (Ours)} & 2026 &
\textbf{0.7703}$_{\text{+34.9\%}}$ &
\textbf{0.6955}$_{\text{+36.0\%}}$ &
\textbf{0.0741}$_{\text{-31.9\%}}$ &
\textbf{0.8603}$_{\text{+14.4\%}}$ &
\textbf{0.8044}$_{\text{+15.4\%}}$ &
\textbf{0.7102}$_{\text{+20.7\%}}$ &
\textbf{0.6290}$_{\text{+19.0\%}}$ &
\textbf{0.0858}$_{\text{-17.8\%}}$ &
\textbf{0.8276}$_{\text{+8.7\%}}$ &
\textbf{0.7652}$_{\text{+8.1\%}}$ \\
\bottomrule
\end{tabular}
}
\vspace{-1mm}
\end{table*}

\begin{table*}[t]
\centering
\caption{
Performance across different retrieval subsets on Test-Base and Test-Novel.
The $x\text{p}y\text{n}$ configuration indicates setups with $x$ positive and $y$ negative objects. \textbf{Bold} marks the best results, and \underline{underline} denotes the second-best.}
\vspace{-2mm}
\label{tab:category_dice_combined}
\setlength{\tabcolsep}{4.5pt}
\renewcommand{\arraystretch}{1.1}
\resizebox{\linewidth}{!}{
\begin{tabular}{l|c|ccccccc|ccccccc}
\toprule
\multirow{2}{*}{Method} & \multirow{2}{*}{Year} & \multicolumn{7}{c|}{\dataset{}-Test-Base} & \multicolumn{7}{c}{\dataset{}-Test-Novel} \\
& & All & 1p0n & 1p1n & 1p2n & 2p0n & 2p1n & 3p0n
& All & 1p0n & 1p1n & 1p2n & 2p0n & 2p1n & 3p0n \\
\midrule
CLIP4CIR~\cite{baldrati2023composed} & 2023 & 0.5333 & 0.6637 & 0.4828 & 0.5209 & 0.4428 & 0.3035 & 0.3802 & 0.5420 & 0.6447 & 0.4595 & 0.4990 & 0.3371 & 0.2494 & 0.2889 \\
BLIP4CIR~\cite{liu2023candidate} & 2023 & 0.5146 & 0.6359 & 0.4732 & 0.5001 & 0.4324 & 0.2908 & 0.3511 & 0.5032 & 0.5825 & 0.4505 & 0.4455 & 0.3441 & 0.2577 & 0.2810 \\
BLIP24CIR~\cite{xu2024sentence} & 2024 & 0.5157 & 0.6351 & 0.4770 & 0.4978 & 0.4230 & 0.3014 & 0.3824 & 0.5097 & 0.6162 & 0.4029 & 0.4382 & 0.3320 & 0.2383 & 0.2301 \\
Bi-BLIP4CIR~\cite{liu2024bi} & 2024 & 0.5308 & 0.6622 & 0.4855 & 0.5023 & 0.4361 & 0.3061 & 0.3666 & 0.5490 & 0.6364 & 0.4796 & 0.5377 & 0.3688 & \underline{0.2937} & \underline{0.3026} \\
CLIP4CIR-SPN~\cite{feng2024improving} & 2024 & 0.5474 & 0.6738 & 0.5034 & 0.5439 & 0.4514 & 0.3223 & 0.3898 & 0.5545 & 0.6525 & 0.4817 & 0.5224 & 0.3522 & 0.2603 & 0.2913 \\
BLIP4CIR-SPN~\cite{feng2024improving} & 2024 & 0.5277 & 0.6547 & 0.4828 & 0.5271 & 0.4354 & 0.2958 & 0.3603 & 0.5222 & 0.6047 & 0.4670 & 0.4735 & 0.3529 & 0.2671 & 0.2909 \\
BLIP24CIR-SPN~\cite{feng2024improving} & 2024 & 0.5709 & \underline{0.6983} & 0.5454 & \underline{0.5531} & 0.4601 & 0.3295 & 0.3961 & \underline{0.5883} & \underline{0.6969} & 0.4846 & \underline{0.5443} & \underline{0.3968} & 0.2847 & 0.2758 \\
CompoDiff~\cite{gu2023compodiff} & 2024 & 0.5446 & 0.6760 & 0.5011 & 0.5120 & 0.4489 & 0.3161 & 0.3863 & 0.5501 & 0.6508 & 0.4709 & 0.5008 & 0.3507 & 0.2627 & 0.2877 \\
ENCODER~\cite{li2025encoder} & 2025 & 0.5530 & 0.6846 & 0.5097 & 0.5199 & 0.4568 & 0.3247 & 0.3937 & 0.5592 & 0.6593 & 0.4813 & 0.5074 & 0.3623 & 0.2703 & 0.2893 \\
ConText-CIR~\cite{xing2025context} & 2025 & 0.5615 & 0.6924 & 0.5226 & 0.5350 & 0.4607 & \underline{0.3306} & 0.3956 & 0.5702 & 0.6715 & 0.4886 & 0.5186 & 0.3754 & 0.2765 & 0.2886 \\
FineCIR~\cite{li2025finecir} & 2025 & \underline{0.5712} & \underline{0.6983} & \underline{0.5467} & 0.5519 & \underline{0.4614} & 0.3278 & \underline{0.3967} & 0.5835 & 0.6917 & \underline{0.4918} & 0.5399 & 0.3694 & 0.2847 & 0.2748 \\
\rowcolor{c1!30}
\textbf{\model{} (Ours)} & 2026 &
\textbf{0.7703} & \textbf{0.8644} & \textbf{0.6335} & \textbf{0.7165} & \textbf{0.8465} & \textbf{0.6210} & \textbf{0.7744}
& \textbf{0.7102} & \textbf{0.8120} & \textbf{0.5317} & \textbf{0.6840} & \textbf{0.6075} & \textbf{0.4977} & \textbf{0.6170} \\
\bottomrule
\end{tabular}
}
\vspace{-1mm}
\end{table*}

\subsection{Experimental Settings}\label{subsec:experiments_setup}
\minisection{Datasets and Supervision Settings.}
Experiments are conducted on \dataset{}, which is divided into Train, Test-Base, and Test-Novel splits.
The Train split is used to optimize \model{}, Test-Base evaluates categories covered by the training taxonomy, and Test-Novel evaluates held-out categories for category-level generalization.
Each training sample provides a composed expression consisting of a reference image $I_{ref}$, a reference object mask $M_{ref}$, and retrieval text $T_{ret}$, together with a target image $I_{tar}$ and its target object mask.
The model uses the composed expression and target image as inputs, while the target mask supervises segmentation and region-level alignment.
Category labels are not used as model inputs.

\minisection{Evaluation Metrics.}
We report Dice, IoU, mDice, mIoU, and MAE.
Higher Dice, IoU, mDice, and mIoU indicate better object retrieval and localization, while lower MAE indicates more accurate mask prediction.

\minisection{Comparable Methods.}
We compare CIR methods on \dataset{} using a \textit{detection model + CIR model + segmentation model} pipeline.
We adopt Detic~\cite{zhou2022detecting} as the detection model, which is pre-trained on LVIS~\cite{gupta2019lvis}.
For segmentation, we employ SAM~\cite{kirillov2023segment}.
For retrieval, we integrate existing CIR models that compute feature similarity between a reference input and candidate regions.
The baseline pipeline proceeds as follows:
1) Detic identifies up to 30 candidate objects in the target image (confidence $> 0.3$, NMS threshold $< 0.8$);
2) Candidate regions are extracted as cropped patches using detected bounding boxes;
3) CIR models compute feature similarity between the reference object and each candidate, selecting the most similar region;
4) The selected bounding box and target image are fed into SAM to produce the target object mask.
This modular pipeline provides a competitive baseline for evaluating \model{} on the \task{} task.

\begin{figure*}[ht]
    \centering
    \includegraphics[width=\linewidth]{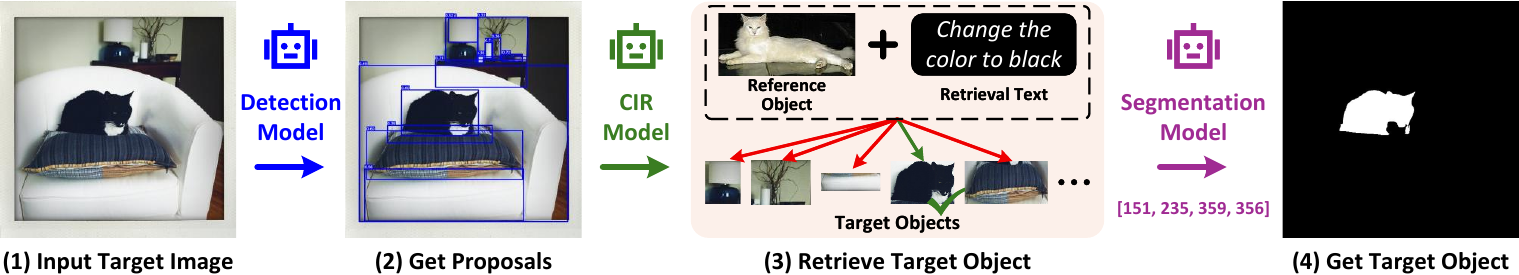}
    \vspace{-6mm}
    \caption{
    \textbf{The modular baseline adapts image-level CIR to object-level mask prediction.} The pipeline combines the \textbf{Detection Model}, \textbf{CIR Model}, and \textbf{Segmentation Model} for \task{}. Given a reference image, object mask, retrieval text, and target image, the detection model generates candidate boxes. The CIR model selects the region best matching the composed expression, and the segmentation model predicts its mask. Errors may accumulate across the pipeline.
    }
     \vspace{-4mm}
    \label{fig:pipeline_compare_method}
\end{figure*}

The existing CIR models cannot be directly applied to the \task{} task. To adapt these models for \task{} and enable comparison with our proposed method, we design a \textit{``Detection Model + CIR Model + Segmentation Model''} pipeline, as illustrated in \cref{fig:pipeline_compare_method}. Based on this pipeline, composed retrieval can be performed on a target image, producing the mask of the retrieved target object.
For the CIR models, we adopt the following implementations: CLIP4CIR~\cite{baldrati2023composed}\footnote{\url{https://github.com/ABaldrati/CLIP4Cir}}, BLIP4CIR~\cite{liu2023candidate}\footnote{\url{https://github.com/Cuberick-Orion/Candidate-Reranking-CIR}}, BLIP24CIR~\cite{xu2024sentence}\footnote{\url{https://github.com/chunmeifeng/SPRC}}, CLIP4CIR-SPN~\cite{feng2024improving}\footnote{\url{https://github.com/BUAADreamer/SPN4CIR}}, BLIP4CIR-SPN~\cite{feng2024improving}\footnote{\url{https://github.com/BUAADreamer/SPN4CIR}}, BLIP24CIR-SPN~\cite{feng2024improving}\footnote{\url{https://github.com/BUAADreamer/SPN4CIR}}, Bi-BLIP4CIR~\cite{liu2024bi}\footnote{\url{https://github.com/Cuberick-Orion/Bi-Blip4CIR}}, CompoDiff~\cite{gu2023compodiff}\footnote{\url{https://github.com/navervision/CompoDiff}}, ENCODER~\cite{li2025encoder}\footnote{\url{https://github.com/JackyLiuAI/ENCODER}}, ConText-CIR~\cite{xing2025context}\footnote{\url{https://github.com/mvrl/ConText-CIR}}, and FineCIR~\cite{li2025finecir}\footnote{\url{https://github.com/SDU-L/FineCIR}}.
These are existing open-source works, most of which provide pre-trained weights for direct evaluation. For models without publicly available weights, we re-train them to ensure fair comparison. Besides, we adopt Detic~\cite{zhou2022detecting}\footnote{\url{https://github.com/facebookresearch/Detic}} as the detection model, pre-trained on LVIS~\cite{gupta2019lvis}. For segmentation, we employ SAM~\cite{kirillov2023segment}\footnote{\url{https://github.com/facebookresearch/segment-anything}}.
Since existing CIR models are not originally designed for object-level mask prediction, the modular pipeline evaluates whether image-level composed retrieval representations can be adapted to \task{} through strong detection and segmentation components.
We include an LLM-assisted baseline, Qwen2.5-VL-7B + SAM, where the LLM predicts a target box from the composed expression and SAM converts the box into a mask.
This baseline tests if direct VLM prompting can replace task-specific composed object alignment.

\begin{figure*}
\centering
\includegraphics[width=\linewidth]{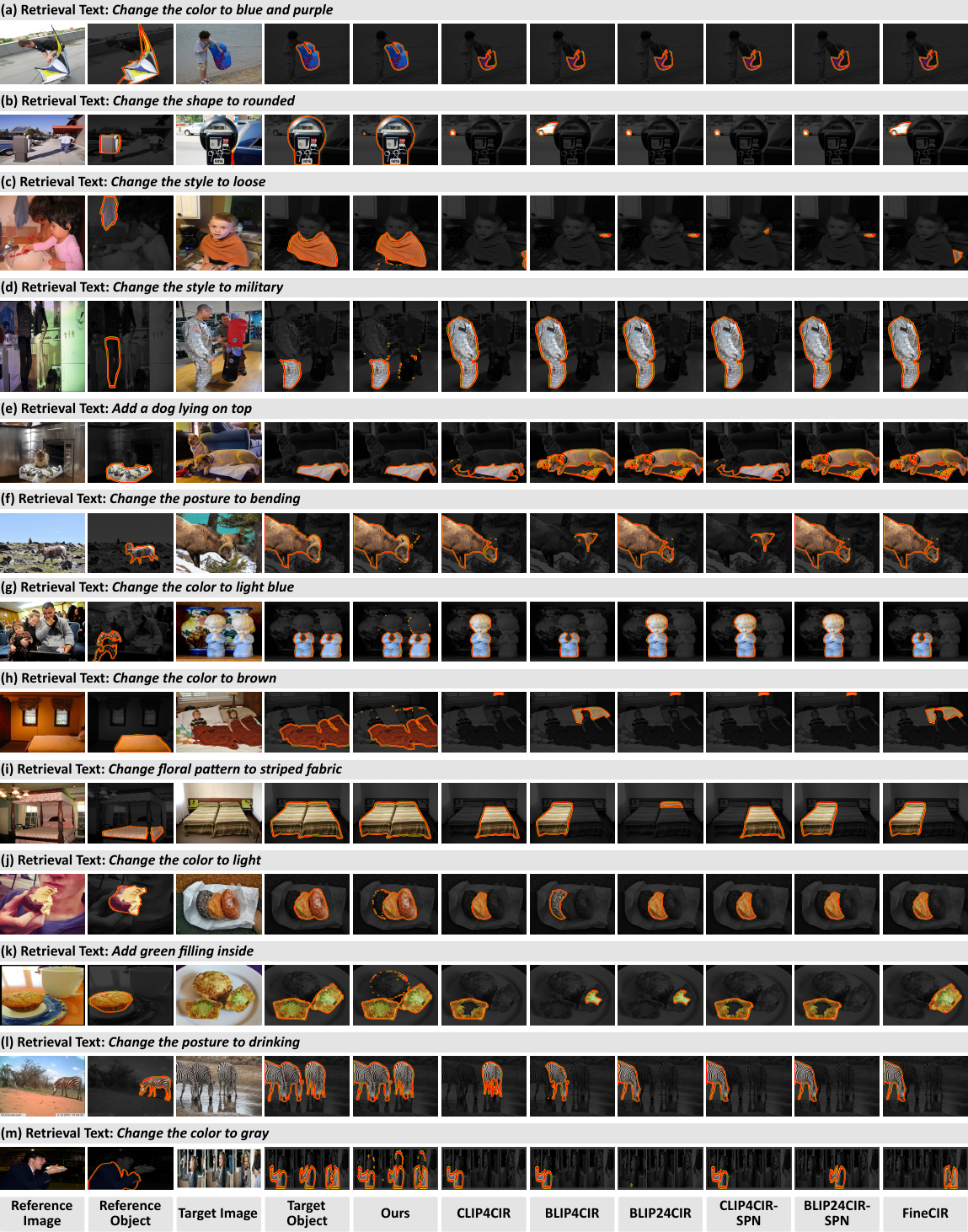}
\vspace{-6mm}
\caption{\textbf{\model{} produces more accurate object-level retrieval masks.}
From left to right: (1) Reference Image $I_{ref}$; (2) Reference Object $O_{ref}$; (3) Target Image $I_{tar}$; (4) Target Object $O_{tar}$; (5) Ours; (6) CLIP4CIR; (7) BLIP4CIR; (8) BLIP24CIR; (9) CLIP4CIR-SPN; (10) BLIP24CIR-SPN; (11) FineCIR.
}
\vspace{-4mm}
\label{fig:experiment_visual}
\end{figure*}

\subsection{Quantitative Results}
We evaluate \model{} on \dataset{} with 11 strong CIR baselines for comparison: CLIP4CIR~\cite{baldrati2023composed}, BLIP4CIR~\cite{liu2023candidate}, BLIP24CIR~\cite{xu2024sentence}, CLIP4CIR-SPN~\cite{feng2024improving}, BLIP4CIR-SPN~\cite{feng2024improving}, BLIP24CIR-SPN~\cite{feng2024improving}, Bi-BLIP4CIR~\cite{liu2024bi}, CompoDiff~\cite{gu2023compodiff}, ENCODER~\cite{li2025encoder}, ConText-CIR~\cite{xing2025context}, FineCIR~\cite{li2025finecir}.
For CompoDiff, ConText-CIR, and FineCIR, since pre-trained weights are unavailable, we reproduced their results by retraining each model using the official implementations.
All other baselines use publicly released weights pre-trained on the CIRR~\cite{liu2021image}.

Results are summarized in \cref{tab:baseline}.
On Test-Base, \model{} surpasses the second-best method, achieving substantial improvements of 34.9\% in Dice, 36.0\% in IoU, 14.4\% in mDice, and 15.4\% in mIoU. On Test-Novel, it records gains of 20.7\%, 19.0\%, 8.7\%, and 8.1\%, respectively.
These improvements stem from \model{}'s unified end-to-end design, which seamlessly integrates reference feature encoding, adaptive vision-language interaction, and region-level contrastive learning.

\begin{table*}[ht]
\centering
\caption{Performance comparison across head (top 25\%), middle (middle 50\%), and tail (bottom 25\%) category groups on \textbf{\dataset{}-Test-Base}. \textbf{Bold} indicates the best result, and \underline{underline} denotes the second-best.}
\vspace{-2mm}
\label{tab:longtail}
\setlength{\tabcolsep}{11pt}
\renewcommand{\arraystretch}{0.95}
\resizebox{\linewidth}{!}{
\begin{tabular}{l | ccc | ccc | ccc}
\toprule
\multirow{3}{*}{Method} & \multicolumn{9}{c}{\textbf{\dataset{}-Test-Base}} \\
\cmidrule(lr){2-10}
 & \multicolumn{3}{c|}{Head Category (Top 25\%)} & \multicolumn{3}{c|}{Middle Category (Middle 50\%)} & \multicolumn{3}{c}{Tail Category (Bottom 25\%)} \\
 & Dice $\uparrow$ & IoU $\uparrow$ & MAE $\downarrow$ & Dice $\uparrow$ & IoU $\uparrow$ & MAE $\downarrow$ & Dice $\uparrow$ & IoU $\uparrow$ & MAE $\downarrow$ \\
\midrule
CLIP4CIR~\cite{baldrati2023composed} & 0.5535 & 0.4945 & 0.1192 & 0.4171 & 0.3766 & 0.1025 & 0.4110 & 0.3789 & 0.0892 \\
BLIP4CIR~\cite{liu2023candidate} & 0.5261 & 0.4664 & 0.1288 & 0.4448 & 0.3994 & 0.1043 & \underline{0.5246} & \underline{0.4829} & 0.0854 \\
BLIP24CIR~\cite{xu2024sentence} & 0.5262 & 0.4670 & 0.1221 & 0.4535 & 0.4078 & \underline{0.0947} & 0.4805 & 0.4335 & \underline{0.0740} \\
Bi-BLIP4CIR~\cite{liu2024bi} & 0.5437 & 0.4838 & 0.1264 & 0.4506 & 0.4045 & 0.1052 & \textbf{0.5692} & \textbf{0.5257} & \textbf{0.0715} \\
CLIP4CIR-SPN~\cite{feng2024improving} & 0.5670 & 0.5063 & 0.1162 & 0.4346 & 0.3927 & 0.1001 & 0.4252 & 0.3932 & 0.0825 \\
BLIP4CIR-SPN~\cite{feng2024improving} & 0.5397 & 0.4791 & 0.1268 & 0.4546 & 0.4084 & 0.1030 & 0.5259 & 0.4843 & 0.0857 \\
BLIP24CIR-SPN~\cite{feng2024improving} & \underline{0.5819} & \underline{0.5182} & \underline{0.1131} & \underline{0.5082} & \underline{0.4589} & \textbf{0.0906} & 0.4877 & 0.4437 & 0.0825 \\
\rowcolor{c1!30}
\textbf{\model{} (Ours)} & \textbf{0.7984} & \textbf{0.7254} & \textbf{0.0695} & \textbf{0.6100} & \textbf{0.5245} & 0.1006 & 0.5616 & 0.4818 & 0.1028 \\
\bottomrule
\end{tabular}
}
\vspace{-2mm}
\end{table*}

\begin{table*}[ht]
\centering
\caption{Performance comparison across head (top 25\%), middle (middle 50\%), and tail (bottom 25\%) category groups on \textbf{\dataset{}-Test-Novel}. \textbf{Bold} indicates the best result, and \underline{underline} denotes the second-best.}
\vspace{-2mm}
\label{tab:longtail_novel}
\setlength{\tabcolsep}{11pt}
\renewcommand{\arraystretch}{0.95}
\resizebox{\linewidth}{!}{
\begin{tabular}{l | ccc | ccc | ccc}
\toprule
\multirow{3}{*}{Method} & \multicolumn{9}{c}{\textbf{\dataset{}-Test-Novel}} \\
\cmidrule(lr){2-10}
 & \multicolumn{3}{c|}{Head Category (Top 25\%)} & \multicolumn{3}{c|}{Middle Category (Middle 50\%)} & \multicolumn{3}{c}{Tail Category (Bottom 25\%)} \\
 & Dice $\uparrow$ & IoU $\uparrow$ & MAE $\downarrow$ & Dice $\uparrow$ & IoU $\uparrow$ & MAE $\downarrow$ & Dice $\uparrow$ & IoU $\uparrow$ & MAE $\downarrow$ \\
\midrule
CLIP4CIR~\cite{baldrati2023composed} & 0.5645 & 0.5108 & 0.1165 & 0.4625 & 0.4185 & 0.1100 & 0.3842 & 0.3386 & 0.0857 \\
BLIP4CIR~\cite{liu2023candidate} & 0.5113 & 0.4520 & 0.1340 & 0.4715 & 0.4235 & 0.1202 & \underline{0.5045} & \underline{0.4463} & 0.0756 \\
BLIP24CIR~\cite{xu2024sentence} & 0.5190 & 0.4620 & 0.1214 & 0.4772 & 0.4293 & 0.1068 & 0.4440 & 0.3890 & \underline{0.0741} \\
Bi-BLIP4CIR~\cite{liu2024bi} & 0.5676 & 0.5077 & 0.1215 & 0.4812 & 0.4332 & 0.1194 & 0.4612 & 0.4070 & 0.0882 \\
CLIP4CIR-SPN~\cite{feng2024improving} & 0.5742 & 0.5179 & 0.1135 & 0.4862 & 0.4398 & 0.1077 & 0.3987 & 0.3538 & 0.0833 \\
BLIP4CIR-SPN~\cite{feng2024improving} & 0.5315 & 0.4715 & 0.1296 & 0.4866 & 0.4376 & 0.1196 & 0.5042 & 0.4455 & 0.0742 \\
BLIP24CIR-SPN~\cite{feng2024improving} & \underline{0.6030} & \underline{0.5417} & \underline{0.1060} & \underline{0.5345} & \underline{0.4810} & \underline{0.1002} & \textbf{0.5176} & \textbf{0.4547} & \textbf{0.0684} \\
\rowcolor{c1!30}
\textbf{\model{} (Ours)} & \textbf{0.7353} & \textbf{0.6527} & \textbf{0.0827} & \textbf{0.6244} & \textbf{0.5494} & \textbf{0.0966} & 0.4791 & 0.3881 & 0.1102 \\
\bottomrule
\end{tabular}
}
\vspace{-4mm}
\end{table*}

\minisection{Performance by Category Frequency.}
As shown in \cref{fig:dataset_histogram}, categories are ranked by retrieval-triplet frequency and separated into shaded head, middle, and tail regions.
To analyze model robustness under this long-tailed structure, we divide categories by rank: the top 25\% are defined as \textbf{head categories}, the middle 50\% as \textbf{middle categories}, and the bottom 25\% as \textbf{tail categories}.
\cref{fig:dataset_histogram_2} shows the resulting category and sample mass within Test-Base and Test-Novel, highlighting that the head group occupies a small category fraction but contributes most samples, while tail categories are numerous but sparsely sampled.
In Test-Base, the Head, Middle, and Tail groups contain 71, 142, and 71 categories, with 19,897, 3,282, and 158 samples, respectively.
In Test-Novel, the Head, Middle, and Tail groups include 19, 39, and 20 categories, with 14,101, 3,605, and 195 samples.
This imbalance makes tail-category scores more sensitive to individual samples, so we interpret head, middle, and tail results jointly rather than in isolation.
As summarized in \cref{tab:longtail,tab:longtail_novel}, our method (\model{}) consistently outperforms existing baselines across most category groups.
In Test-Base, it achieves a Dice of 0.7984 on head categories, surpassing the best baseline (BLIP24CIR-SPN, 0.5819) by a relative gain of +37.2\%, and an IoU of 0.7254 with a relative gain of +40.0\%.
For middle categories, \model{} also leads significantly (0.6100 vs. 0.5082 Dice, +20.0\% relative gain).
In Test-Novel, \model{} maintains strong results with a head-category Dice of 0.7353, outperforming BLIP24CIR-SPN (0.6030) by a relative gain of +21.9\%.
For middle categories, it achieves 0.6244 Dice, exceeding the baseline by a relative gain of +16.8\%.
Even on tail categories, where samples are few, performance remains on par with top-performing methods.

\begin{table}
\centering
\caption{Comparison with an LLM-assisted prompting baseline.}
\vspace{-2mm}
\label{tab:llm_sam_baseline}
\setlength{\tabcolsep}{4pt}
\renewcommand{\arraystretch}{0.95}
\resizebox{\linewidth}{!}{
\begin{tabular}{l|cc|cc}
\toprule
\multirow{2}{*}{Method} & \multicolumn{2}{c|}{\dataset{}-Test-Base} & \multicolumn{2}{c}{\dataset{}-Test-Novel} \\
& Dice $\uparrow$ & IoU $\uparrow$ & Dice $\uparrow$ & IoU $\uparrow$ \\
\midrule
Qwen2.5-VL-7B + SAM & 0.5710 & 0.5105 & 0.5714 & 0.5110 \\
\rowcolor{c1!30}
\model{} (Ours) & \textbf{0.7703} & \textbf{0.6955} & \textbf{0.7102} & \textbf{0.6290} \\
\bottomrule
\end{tabular}}
\vspace{-6mm}
\end{table}

\minisection{LLM-Assisted Baseline.}
We compare with Qwen2.5-VL-7B + SAM in \cref{tab:llm_sam_baseline} to evaluate whether an instruction-following VLM can directly infer target boxes from the composed expression.
This baseline is strong because it leverages open-world visual reasoning and delegates mask generation to SAM.
However, it remains a multi-stage solution and can suffer from box ambiguity when multiple same-category or semantically related objects satisfy part of the composed expression.

\begin{figure}[t]
\vspace{-2mm}
\centering
\includegraphics[width=\linewidth]{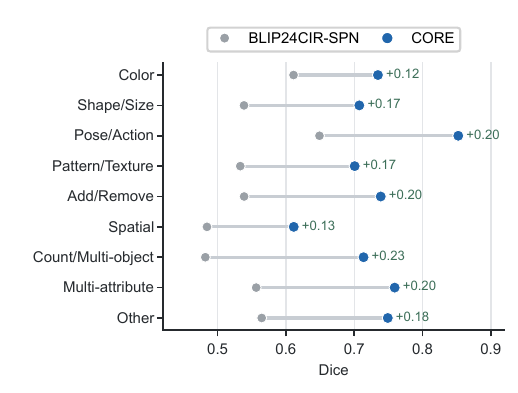}
\vspace{-10mm}
\caption{\textbf{\model{} improves across retrieval-text types.}
Dice performance is shown for keyword-based operation groups. Each line connects BLIP24CIR-SPN and \model{}, and the green label reports the absolute Dice gain of \model{}.
}
\vspace{-6mm}
\label{fig:text_attribute_type_performance}
\end{figure}

\minisection{Performance Across Compositional Text Types.}
To further examine where \model{} improves over retrieval-segmentation baselines, we group test triplets according to the dominant operation described in the retrieval text.
The groups include color, shape/size, pose/action, pattern/texture, add/remove, spatial relation, count or multi-object expressions, multi-attribute expressions, and other descriptions.
This diagnostic uses keyword-based assignment, so the groups should be interpreted as an approximate operation-level breakdown rather than a manually verified linguistic taxonomy.
As shown in \cref{fig:text_attribute_type_performance}, \model{} consistently outperforms BLIP24CIR-SPN across all text-operation groups.
The improvement is not limited to common color edits: \model{} improves color cases from 0.611 to 0.735 Dice over 15,285 triplets, giving a +0.124 absolute gain.
Larger gains appear for more compositional instructions, including pose/action (+0.203), add/remove (+0.200), and multi-attribute descriptions (+0.203).
The count/multi-object group is small with 54 triplets, but its +0.232 gain further suggests that object-aware composition is valuable when retrieval text specifies multiple valid targets.

\begin{figure}[t]
\centering
\includegraphics[width=\linewidth]{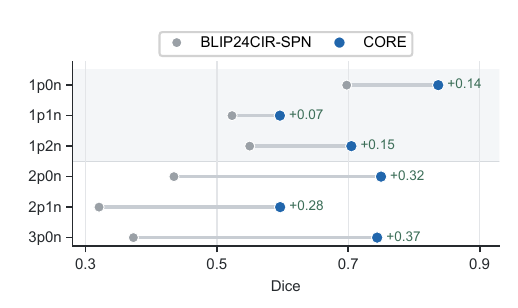}
\vspace{-10mm}
\caption{\textbf{\model{} is robust to multi-target retrieval.}
Dice performance is shown under different target settings, where $x\text{p}y\text{n}$ denotes \textit{x} positive objects and \textit{y} annotated negative objects. Each line connects BLIP24CIR-SPN and \model{}, and the green label reports the absolute Dice gain of \model{}.
}
\vspace{-4mm}
\label{fig:distractor_count_performance}
\end{figure}

\minisection{Robustness to Distractors and Multiple Target Objects.}
We further examine how target-set complexity affects retrieval segmentation over all test triplets.
The $x\text{p}y\text{n}$ setting denotes retrieval with \textit{x} positive target objects to segment and \textit{y} annotated negative objects to reject.
As shown in \cref{fig:distractor_count_performance}, \model{} improves over BLIP24CIR-SPN in every setting.
The gain is +0.139 Dice for the simple 1p0n case and remains positive when distractors are introduced, including 1p1n (+0.073) and 1p2n (+0.155).
The advantage becomes larger in multi-target settings, reaching +0.315 on 2p0n, +0.276 on 2p1n, and +0.371 on 3p0n.
These results indicate that \model{} is especially effective when the output mask must cover multiple composed targets while suppressing irrelevant objects.

The split-wise results in \cref{tab:category_dice_combined} show a consistent pattern on Test-Base and Test-Novel.
On Test-Base, \model{} achieves significant Dice improvements: 1p0n (+23.8\%), 1p1n (+15.8\%), 1p2n (+29.5\%), 2p0n (+83.4\%), 2p1n (+87.8\%), and 3p0n (+95.2\%).
On Test-Novel, gains remain consistent: 1p0n (+16.5\%), 1p1n (+8.1\%), 1p2n (+25.7\%), 2p0n (+53.1\%), 2p1n (+69.46\%), and 3p0n (+103.9\%).
Notably, negative object interference (\eg, 1p1n, 1p2n, 2p1n) increases retrieval complexity, degrading performance across all models.
Settings with three or more positive objects (\eg, 3p0n) are particularly challenging due to heightened semantic ambiguity, yet \model{} maintains robust performance, demonstrating its strength in multi-object retrieval and negative object discrimination.
\begin{table}[t]
\centering
\caption{False positive rate on negative objects. Lower is better.}
\vspace{-2mm}
\label{tab:fpr}
\setlength{\tabcolsep}{18pt}
\renewcommand{\arraystretch}{0.95}
\resizebox{\linewidth}{!}{
\begin{tabular}{l|cc}
\toprule
Method & Test-Base $\downarrow$ & Test-Novel $\downarrow$ \\
\midrule
CLIP4CIR & 0.35 & 0.34 \\
BLIP24CIR & 0.32 & 0.31 \\
BLIP24CIR-SPN & 0.30 & 0.28 \\
ENCODER & 0.27 & 0.29 \\
ConText-CIR & 0.26 & 0.26 \\
FineCIR & 0.25 & 0.24 \\
\rowcolor{c1!30}
\model{} (Ours) & \textbf{0.12} & \textbf{0.15} \\
\bottomrule
\end{tabular}
}
\vspace{-2mm}
\end{table}

\minisection{Negative Object Filtering.}
To directly evaluate distractor suppression, we compute the false positive rate (FPR) over annotated non-target objects.
Target objects are treated as positives, while other annotated objects in the target image are treated as negatives.
A prediction is counted as a false positive if it overlaps with a negative object with IoU $\geq 0.5$.
As shown in \cref{tab:fpr}, this analysis directly measures whether a model incorrectly activates visually similar distractors rather than only evaluating overlap with the target mask.

\begin{table}[t]
\centering
\caption{Inference efficiency comparison.}
\vspace{-2mm}
\label{tab:efficiency}
\setlength{\tabcolsep}{11pt}
\renewcommand{\arraystretch}{0.95}
\resizebox{\linewidth}{!}{
\begin{tabular}{l|ccc}
\toprule
Method & FPS $\uparrow$ & Params (M) $\downarrow$ & FLOPs (G) $\downarrow$ \\
\midrule
CLIP4CIR & 6.1 & 315.22 & 690.35 \\
BLIP4CIR & 5.5 & 360.66 & 755.40 \\
BLIP24CIR & 3.3 & 1496.02 & 1267.62 \\
Bi-BLIP4CIR & 5.9 & 360.66 & 747.17 \\
FineCIR & 2.9 & 1694.92 & 1352.00 \\
\rowcolor{c1!30}
\model{} (Ours) & \textbf{7.2} & \textbf{322.00} & \textbf{491.21} \\
\bottomrule
\end{tabular}}
\vspace{-4mm}
\end{table}

\minisection{Inference Efficiency.}
We further compare inference efficiency with representative modular baselines in \cref{tab:efficiency}.
Pipeline methods require object proposal extraction and repeated feature computation for candidate regions, so their cost increases with the number of detected objects.
In contrast, \model{} performs end-to-end dense prediction in a single forward pass, avoiding repeated proposal-level retrieval and segmentation.

\subsection{Qualitative Results}
We show qualitative results across examples \textbf{a}-\textbf{m} in \cref{fig:experiment_visual}, covering three representative challenges in \task{}: retrieving hard-to-name objects, suppressing same-category distractors, and locating multiple target objects from a composed expression. Examples \textbf{a}-\textbf{d} highlight the role of the reference mask in specifying object identity. In \textbf{b}, the retrieval text asks to \textit{``Change the shape to rounded''}; baselines are distracted by visually related circular regions, whereas \model{} preserves the reference identity and retrieves the intended target. Example \textbf{d} further shows fine-grained style composition, including military-style and loose-fit clothing. Examples \textbf{e} and \textbf{f} contain same-category distractors: in \textbf{e}, the retrieval text \textit{``Add a dog lying on top''} is used to retrieve the \textit{mat}, but several baselines activate the dog or nearby same-category regions, matching the salient text-mentioned object rather than the modified object. Examples \textbf{g}-\textbf{m} evaluate multi-object retrieval, where the model must return all valid instances instead of a single region. Same-category negatives make this harder: in \textbf{j}, \textit{``Change the color to light''} can cause baselines to retrieve black desserts, while in \textbf{k}, \textit{``Add green filling inside''} often leads them to retrieve only one valid object. Overall, existing CIR pipelines depend on separate detection and segmentation modules and lack end-to-end pixel-level semantic reasoning, whereas \model{} jointly uses fine-grained semantics and pixel cues to retrieve complete targets and suppress visually similar non-target objects.

\subsection{Ablation Studies}
We perform ablation studies on network modules, loss functions, pre-trained model scaling and components of composed expressions, following the settings in \cref{subsec:experiments_setup}.

\begin{table}[ht]
\centering
\vspace{-2mm}
\caption{Ablation study on network modules and loss function.}
\vspace{-2mm}
\label{tab:ablation_components}
\setlength{\tabcolsep}{3pt}
\renewcommand{\arraystretch}{0.95}
\resizebox{\linewidth}{!}{
\begin{tabular}{l|ccc|ccc}
\toprule
\multirow{2}{*}{Setting} & \multicolumn{3}{c|}{\dataset{}-Test-Base} & \multicolumn{3}{c}{\dataset{}-Test-Novel} \\
& Dice $\uparrow$ & IoU $\uparrow$ & MAE $\downarrow$ & Dice $\uparrow$ & IoU $\uparrow$ & MAE $\downarrow$ \\
\midrule
\circlednum{1} AVTI + $\mathcal{L}_{cor}$ & 0.7638 & 0.6848 & 0.0766 & 0.6913 & 0.6067 & 0.0976 \\
\circlednum{2} RRE + $\mathcal{L}_{cor}$ & 0.7541 & 0.6762 & 0.0808 & 0.6686 & 0.5877 & 0.1055 \\
\circlednum{3} RRE + AVTI & 0.7533 & 0.6768 & 0.0784 & 0.6601 & 0.5791 & 0.1041 \\
\rowcolor{c1!30}
\makecell{\circlednum{4} RRE + AVTI + $\mathcal{L}_{cor}$\\(\textbf{Ours})} &
\textbf{0.7703} & \textbf{0.6955} & \textbf{0.0741} &
\textbf{0.7102} & \textbf{0.6290} & \textbf{0.0858} \\
\bottomrule
\end{tabular}
}

\vspace{-4mm}
\end{table}

\begin{table}[ht]
\centering
\caption{Ablation study on the scaling of SigLIP and SAM.}
\vspace{-2mm}
\label{tab:ablation_scaling}
\setlength{\tabcolsep}{2pt}
\renewcommand{\arraystretch}{0.95}
\resizebox{\linewidth}{!}{
\begin{tabular}{l|ccc|ccc}
\toprule
\multirow{2}{*}{Setting} & \multicolumn{3}{c|}{\dataset{}-Test-Base} & \multicolumn{3}{c}{\dataset{}-Test-Novel} \\
& Dice $\uparrow$ & IoU $\uparrow$ & MAE $\downarrow$ & Dice $\uparrow$ & IoU $\uparrow$ & MAE $\downarrow$ \\
\midrule
\circlednum{1} SigLIP (B) + SAM (L) &
0.7784 & 0.7118 & 0.0692 &
0.7008 & \textbf{0.6338} & 0.0882 \\
\circlednum{2} SigLIP (L) + SAM (B) & 0.7741 & 0.6977 & 0.0729 &
0.6828 & 0.5990 & 0.0989 \\
\circlednum{3} SigLIP (L) + SAM (L) &
\textbf{0.7793} & \textbf{0.7127} & \textbf{0.0682} &
0.6938 & 0.6261 & 0.0917 \\
\rowcolor{c1!30}
\makecell{\circlednum{4} SigLIP (B) + SAM (B)\\(\textbf{Ours})} &
0.7703 & 0.6955 & 0.0741 &
\textbf{0.7102} & 0.6290 & \textbf{0.0858} \\
\bottomrule
\end{tabular}
}
\vspace{-4mm}
\end{table}

\begin{table}[ht]
\centering
\caption{Ablation study on components of composed expressions.}
\vspace{-2mm}
\label{tab:ablation_expression_components}
\setlength{\tabcolsep}{3pt}
\renewcommand{\arraystretch}{0.95}
\resizebox{\linewidth}{!}{
\begin{tabular}{l|ccc|ccc}
\toprule
\multirow{2}{*}{Setting} & \multicolumn{3}{c|}{\dataset{}-Test-Base} & \multicolumn{3}{c}{\dataset{}-Test-Novel} \\
& Dice $\uparrow$ & IoU $\uparrow$ & MAE $\downarrow$ & Dice $\uparrow$ & IoU $\uparrow$ & MAE $\downarrow$ \\
\midrule
\circlednum{1} $I_{ref}$ + $M_{ref}$ & 0.7411 & 0.6601 & 0.0833 & 0.6664 & 0.5813 & 0.1054 \\
\circlednum{2} $I_{ref}$ + $T_{ret}$ & 0.7101 & 0.6302 & 0.0943 & 0.6460 & 0.5644 & 0.1122 \\
\circlednum{3} $I_{ref}$ & 0.6770 & 0.5974 & 0.1057 & 0.6137 & 0.5359 & 0.1213 \\
\circlednum{4} $T_{ret}$ & 0.6767 & 0.6002 & 0.1036 & 0.6144 & 0.5363 & 0.1211 \\
\rowcolor{c1!30}
\makecell{\circlednum{5} $I_{ref}$ + $M_{ref}$ + $T_{ret}$\\(\textbf{Ours})} &
\textbf{0.7703} & \textbf{0.6955} & \textbf{0.0741} &
\textbf{0.7102} & \textbf{0.6290} & \textbf{0.0858} \\
\bottomrule
\end{tabular}
}
\vspace{-2mm}
\end{table}

\minisection{Network Modules and Loss Function.}
As shown in \cref{tab:ablation_components}, removing any of the network modules or contrastive loss results in performance degradation.
\circlednum{1} Replacing RRE with masked pooling decreases Dice by 0.84\% on Test-Base and 2.66\% on Test-Novel.
\circlednum{2} Replacing AVTI with a simple sum operation leads to a 5.85\% Dice drop on Test-Novel.
We also compare AVTI with a standard gated fusion alternative, which obtains Dice scores of 0.7612 on Test-Base and 0.6951 on Test-Novel.
AVTI further improves the results to 0.7703 and 0.7102, showing that adaptive modality-specific interaction is more effective than generic gating for composed object retrieval.
\circlednum{3} Removing $\mathcal{L}_{cor}$ causes the largest decline of 7.05\% on Test-Novel.
These results show that RRE strengthens the representation of the reference object by separating it from the background; the AVTI enhances vision-language alignment, enabling more reliable composed expressions; $\mathcal{L}_{cor}$ improves the alignment between the reference and target object representations while suppressing negative objects.

\minisection{Pre-trained Model Scaling.}
As shown in \cref{tab:ablation_scaling}, we evaluate the effect of scaling the SigLIP and SAM backbones, where B and L denote the Base and Large variants, respectively.
\circlednum{1} SigLIP (B) + SAM (L): Replacing SAM-Base with SAM-Large increases Dice on Test-Base from 0.7703 to 0.7784 (+1.05\%) but reduces Dice on Test-Novel from 0.7102 to 0.7008 (–1.32\%).
\circlednum{2} SigLIP (L) + SAM (B): Enlarging SigLIP yields a slight gain on Test-Base (0.7741, +0.49\%) but a notable drop on Test-Novel (0.6828, –3.87\%).
\circlednum{3} SigLIP (L) + SAM (L): Scaling up both backbones further improves Test-Base Dice to 0.7793 (+1.17\%) yet decreases Test-Novel to 0.6938 (–2.30\%).
Overall, larger pre-trained models enhance in-domain accuracy but tend to overfit, degrading cross-domain generalization. Ours configuration thus achieves the best balance between accuracy, robustness, and efficiency.

\minisection{Components of Composed Expressions.}
As shown in \cref{tab:ablation_expression_components}, we analyze how each component of the composed expression affects retrieval.
\circlednum{1} $I_{ref}$ + $M_{ref}$ (w/o $T_{ret}$): Removing the retrieval text causes the Dice score to drop from 0.7703 to 0.7411 (–3.79\%) on Test-Base and from 0.7102 to 0.6664 (–6.17\%) on Test-Novel.
\circlednum{2} $I_{ref}$ + $T_{ret}$ (w/o $M_{ref}$): Excluding the reference object mask decreases Dice from 0.7703 to 0.7101 (–7.82\%) on Test-Base and from 0.7102 to 0.6460 (–9.04\%) on Test-Novel.
\circlednum{3} $I_{ref}$ (w/o $T_{ret}$, $M_{ref}$): Removing both the retrieval text and mask results in a larger decline, with Dice dropping to 0.6770 (–12.11\%) on Test-Base and 0.6137 (–13.59\%) on Test-Novel.
\circlednum{4} $T_{ret}$ (w/o $I_{ref}$, $M_{ref}$): Retaining only the retrieval text yields a similar drop to 0.6767 (–12.15\%) and 0.6144 (–13.49\%) on Test-Base and Test-Novel, respectively.
Overall, using all three components: retrieval text ($T_{ret}$), reference object mask ($M_{ref}$), and reference image ($I_{ref}$) achieves the best results, highlighting their complementary roles in producing robust and accurate retrieval performance.

\begin{table}[t]
\centering
\caption{Reference-mask sensitivity under test-time perturbations.}
\vspace{-2mm}
\label{tab:reference_mask_sensitivity}
\setlength{\tabcolsep}{12pt}
\renewcommand{\arraystretch}{0.95}
\resizebox{\linewidth}{!}{
\begin{tabular}{l|cc|cc}
\toprule
\multirow{2}{*}{Mask} & \multicolumn{2}{c|}{Dice $\uparrow$} & \multicolumn{2}{c}{$\Delta$Dice} \\
 & Base & Novel & Base & Novel \\
\midrule
\rowcolor{c1!30}
Clean & 0.7703 & 0.7102 & +0.0000 & +0.0000 \\
Eroded & 0.7699 & 0.7103 & -0.0003 & +0.0001 \\
Dilated & 0.7704 & 0.7090 & +0.0001 & -0.0012 \\
Box & 0.7568 & 0.6894 & -0.0135 & -0.0208 \\
Shifted & 0.7645 & 0.6982 & -0.0057 & -0.0120 \\
\bottomrule
\end{tabular}}
\vspace{-4mm}
\end{table}

\begin{figure*}[t]
\centering
\includegraphics[width=\linewidth]{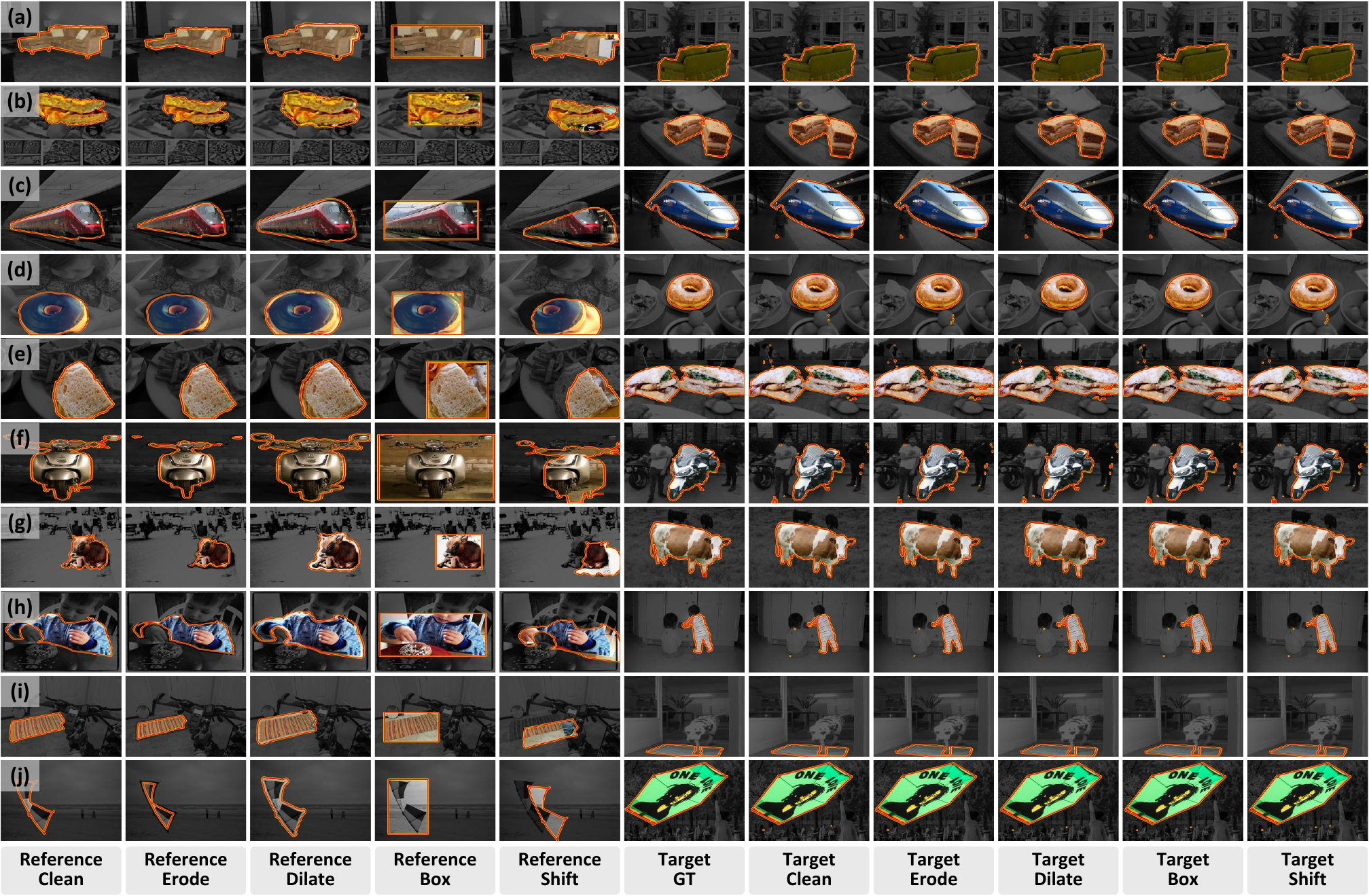}
\vspace{-8mm}
\caption{
\textbf{Reference-mask perturbations preserve most target predictions.}
Examples \textbf{a}-\textbf{j} compare the clean reference mask with eroded, dilated, box-shaped, and shifted variants.
For each case, the target ground truth and predictions under the corresponding perturbations are shown side by side.
Target-side results reveal how reference-mask changes affect composed object retrieval and grounding.
The predictions remain acceptable under different perturbations, indicating the robustness of \model{} to reference-mask noise.
}
\vspace{-4mm}
\label{fig:reference_mask_sensitivity_visual}
\end{figure*}

\begin{figure}[t]
\centering
\includegraphics[width=\linewidth]{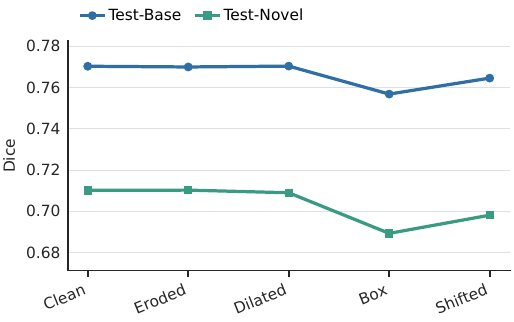}
\vspace{-6mm}
\caption{
\textbf{\model{} remains robust to reference-mask perturbations.}
Absolute Dice is reported for clean, eroded, dilated, box, and shifted reference masks.
}
\vspace{-6mm}
\label{fig:reference_mask_sensitivity_dice}
\end{figure}

\minisection{Sensitivity to Reference-Mask Perturbations.}
We evaluate whether \model{} relies on perfectly annotated reference masks by perturbing $M_{ref}$ at test time while keeping the reference image, target image, retrieval text, model checkpoint, and all other inputs unchanged.
No retraining is performed.
This controlled setting isolates reference-mask quality from retrieval-text ambiguity and target-image difficulty, while reflecting practical cases where masks come from human annotators or automatic segmentation tools.
We consider four perturbations: mask erosion, mask dilation, replacing the mask with its tight axis-aligned bounding box, and spatially shifting the mask with zero padding.
For erosion and dilation, we apply binary morphology with an elliptical structuring element whose side length is the nearest odd integer to $\min(H,W)/80$ and use one iteration.
If erosion removes all foreground pixels, we keep the original mask to avoid invalid inputs.
For the box variant, all pixels inside the tight foreground bounding rectangle are set as foreground.
For the shifted variant, the mask is translated by $d_x=0.04W$ and $d_y=0.025H$, with offsets rounded to integers and empty regions padded with zeros.
These perturbations cover boundary uncertainty, coarse spatial support, and reference-object localization error.

\cref{fig:reference_mask_sensitivity_visual} further visualizes reference-mask perturbations on ten selected examples, ordered as examples \textbf{a}-\textbf{j}.
The retrieval texts for examples \textbf{a}-\textbf{j} are: \textit{``Add a skirt at the bottom''}, \textit{``Change the color to brown''}, \textit{``Change the color to silver-blue''}, \textit{``Change the color to light''}, \textit{``Add lettuce''}, \textit{``Add a windshield''}, \textit{``Change the color to white-brown''}, \textit{``Change the color to striped''}, \textit{``Change the texture to rubber''}, and \textit{``Change the color to green''}, respectively.
These selected examples show that \model{} can preserve the intended target prediction even under visibly eroded, dilated, box-shaped, and shifted reference masks.
As shown in \cref{tab:reference_mask_sensitivity,fig:reference_mask_sensitivity_dice}, boundary perturbations have little effect: erosion and dilation change Dice by at most 0.0012 on both splits.
Box and shift perturbations are more challenging because they include surrounding context or displace the reference support, but the degradation remains modest: replacing $M_{ref}$ with a bounding box drops Dice by 0.0135 on Test-Base and 0.0208 on Test-Novel, while spatial shifts reduce Dice by 0.0057 and 0.0120, respectively.
These limited drops are acceptable without retraining or test-time adaptation, indicating that RRE does not require perfectly aligned reference masks.
Overall, RRE is robust to small boundary noise and remains stable under practical coarse-mask and localization perturbations.

\subsection{Failure Cases}
\cref{fig:failure_cases} presents representative failure cases of \model{} and the strong modular baseline BLIP24CIR-SPN.
The retrieval texts for examples \textbf{a}-\textbf{f} are: \textit{``Add black pants''}, \textit{``Add blue clothing''}, \textit{``Change the posture to lying''}, \textit{``Change the pattern to striped''}, \textit{``Change the color to blue''}, and \textit{``Change the color to white''}, respectively.
These cases reveal three main failure modes.
First, visually salient distractors can dominate the composed expression: in \textbf{a}, \model{} incorrectly retrieves the white bear, and in \textbf{b}, it activates toys outside the blue-clothing region.
BLIP24CIR-SPN is also vulnerable in such cases because its modular pipeline first generates candidate boxes and then reranks cropped regions, so an object with strong color or category cues can be selected even when it is not the intended target.
Second, object identity can be lost when the retrieval text describes a broad pose or appearance change; in \textbf{c}, the model retrieves a non-cat object despite the intended cat reference.
This issue is more pronounced for BLIP24CIR-SPN when the detected candidate region does not preserve the reference-object identity, causing the CIR reranker to overemphasize the text attribute.
Third, some apparent failures expose annotation noise rather than only model errors.
In \textbf{d} and \textbf{e}, \model{} retrieves valid objects, including the blue cup and the white clock, whose masks come from LVIS annotations, but these objects do not fully match the intended composed expressions.
This indicates that \dataset{} still contains noisy or ambiguous samples inherited from automatic construction and LVIS masks, especially when attribute descriptions are short or color/pattern cues are not unique.
These cases suggest that future versions of \dataset{} should further refine mask-text consistency, strengthen ambiguity filtering, and add more human checks for samples with visually similar distractors.
\begin{figure}[!htbp]
 \vspace{-4mm}
  \centering
  \includegraphics[width=\linewidth]{figures/fig_failure_cases.pdf}
  \vspace{-8mm}
  \caption{
  \textbf{Failure cases reveal distractor confusion and annotation noise.} Examples include incorrect retrieval under salient distractors, BLIP24CIR-SPN errors caused by modular candidate reranking.
  } \label{fig:failure_cases}
  \vspace{-6mm}
\end{figure}

\section{Limitations and Future Work}
\minisection{Dataset Scope.}
Although \dataset{} provides a large-scale benchmark for \task{}, it is built on COCO and LVIS and may inherit their category, scene, and geographic biases.
The current benchmark mainly focuses on attribute-level modifications such as color, shape, pose, action, and spatial appearance.
More complex compositional expressions involving long-horizon reasoning, relational changes, viewpoint changes, object addition or removal, and richer multi-object constraints remain important directions for future extensions.

\minisection{Domain Transfer.}
The pipeline assumes object masks or reliable mask generators.
Such masks can be obtained from LVIS annotations, SAM-style prompts, grounding models, or domain-specific segmentation tools, but transferring the pipeline to specialized domains such as medical imaging may require domain-adapted vision-language and segmentation models.
Future COR benchmarks can explore broader domains and more realistic user interactions.

\minisection{Model and Deployment.}
Our model is optimized for \dataset{} and improves task-specific composed object alignment, but may trade off open-vocabulary generalization.
Scaling \task{} to high-resolution or large-candidate scenes may require efficient inference, such as tiling, sparse pruning, or lightweight backbones.
Finally, object-level retrieval can raise privacy and misuse concerns, requiring responsible data governance and deployment constraints.

\section{Conclusion}
We introduce \textbf{Composed Object Retrieval (COR)}, a new task that extends multimodal retrieval from image-level matching to object-level grounding with composed expressions. We present \textbf{\dataset{}}, a large-scale dataset containing 125,541 triplets across 408 categories, with pixel-level masks, multi-object retrieval cases, and visually similar distractors for fine-grained evaluation. We also propose \textbf{\model{}}, an end-to-end framework that integrates reference region encoding, adaptive vision-text interaction, and COR-oriented contrastive learning. Extensive experiments show that \model{} outperforms existing methods across base, novel, long-tailed, and challenging retrieval settings. Together, the task, dataset, and model establish a foundation for studying composed retrieval with explicit object grounding. COR enables fine-grained visual search and advanced image understanding, paving the way for next-generation object-level retrieval.

\section*{Acknowledgment}
Tong Wang and Guanyu Yang are supported by the National Natural Science Foundation of China (T2541064, 82441021) and the Jiangsu Province Cutting-edge Technology R\&D Program (BF2025628). This work is also supported by the Big Data Computing Center of Southeast University.

\bibliographystyle{IEEEtran}
\bibliography{egbib}

\newpage


\begin{IEEEbiography}[{\includegraphics[width=1in,height=1.25in,clip,keepaspectratio]{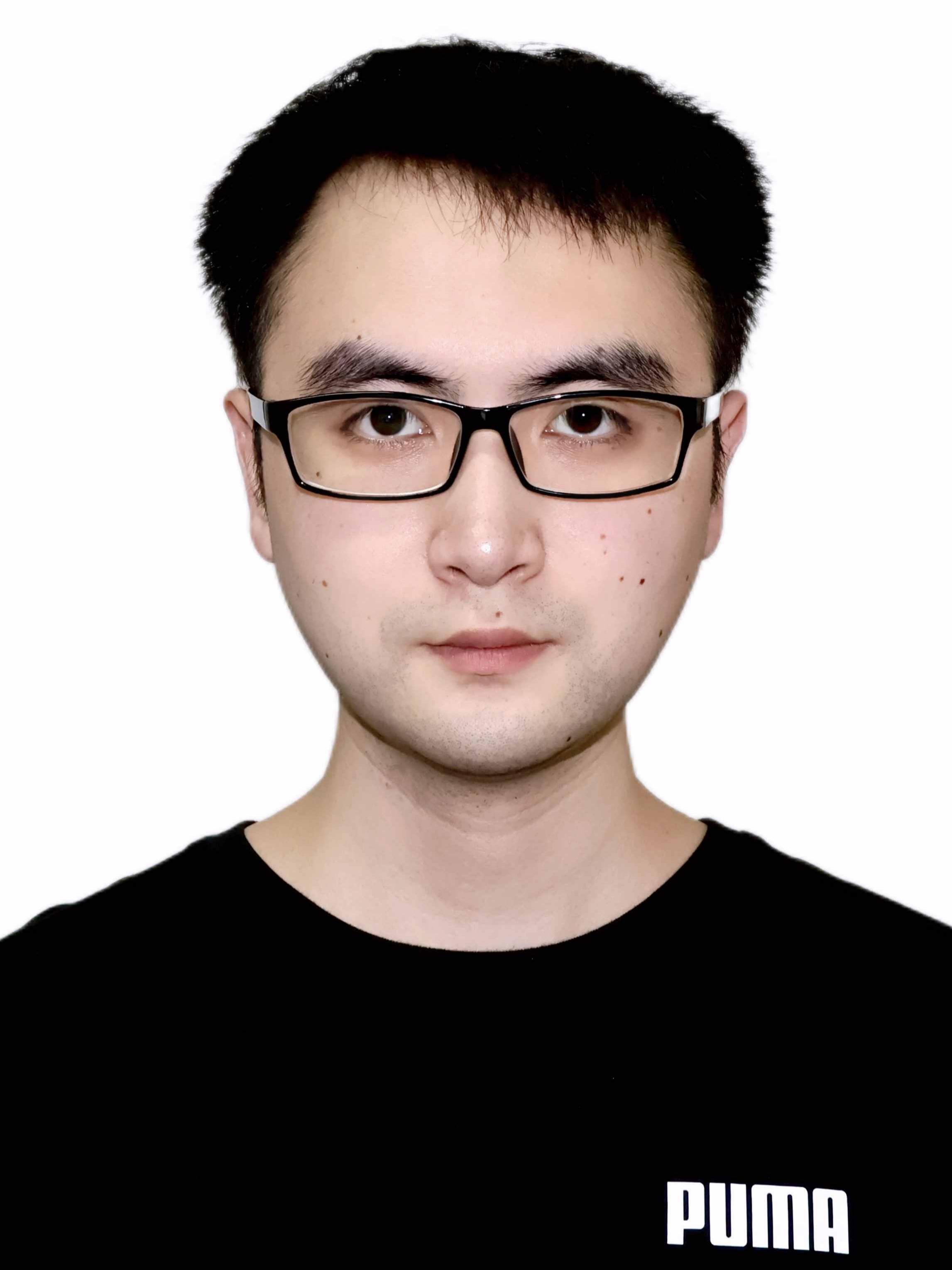}}]{Tong Wang}
is currently a Ph.D. candidate at the Key Laboratory of New Generation Artificial Intelligence Technology and Its Interdisciplinary Applications (Ministry of Education), Southeast University, Nanjing, China. He is also a visiting student at Mohamed bin Zayed University of Artificial Intelligence (MBZUAI). His research interests include medical image analysis and multimodal models.
\end{IEEEbiography}

\vspace{-11mm}

\begin{IEEEbiography}[{\includegraphics[width=1in,height=1.25in,clip,keepaspectratio]{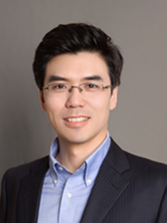}}]{Guanyu Yang}
(Senior Member, IEEE) received the B.S. and M.S. degrees in biomedical engineering from Southeast University, Nanjing, China, in 2002 and 2005, respectively. He joined Southeast University in 2011, where he is currently a Professor and Vice Dean of the School of Computer Science and Engineering. His research interests include medical image analysis, biomedical image processing, and artificial intelligence for healthcare.
\end{IEEEbiography}

\vspace{-11mm}

\begin{IEEEbiography}[{\includegraphics[width=1in,height=1.25in,clip,keepaspectratio]{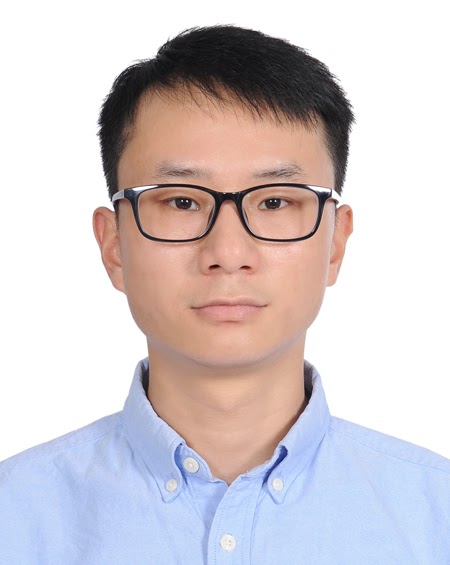}}]{Nian Liu}
(Member, IEEE) received the B.S. and Ph.D. degrees from the School of Automation at Northwestern Polytechnical University, Xi’an, China, in 2012 and 2020, respectively. He was previously a Research Scientist with Mohamed Bin Zayed University of Artificial Intelligence, UAE. He is currently a Professor with the School of Automation, Northwestern Polytechnical University. His research interests include computer vision and deep learning.
\end{IEEEbiography}

\vspace{-11mm}

\begin{IEEEbiography}[{\includegraphics[width=1in,height=1.25in,clip,keepaspectratio]{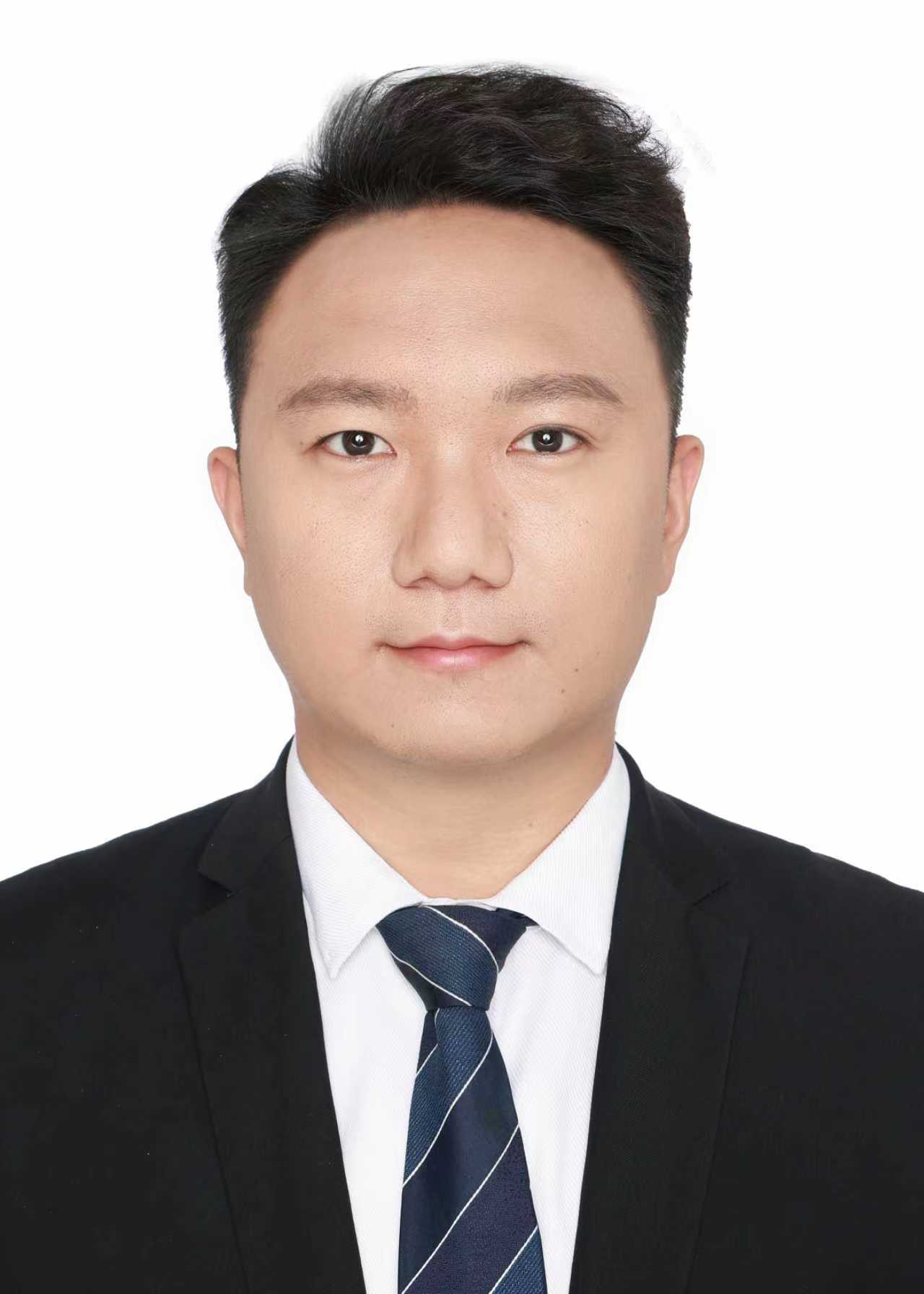}}]{Zongyan Han}
received the B.E. and Ph.D. degrees from Nanjing University of Science and Technology, Nanjing, China. He was a Research Assistant with The Hong Kong Polytechnic University, Hong Kong. He is currently a Postdoctoral Researcher with the Department of Computer Vision, Mohamed bin Zayed University of Artificial Intelligence. His research interests include computer vision and deep learning.
\end{IEEEbiography}

\vspace{-11mm}

\begin{IEEEbiography}[{\includegraphics[width=1in,height=1.25in,clip,keepaspectratio]{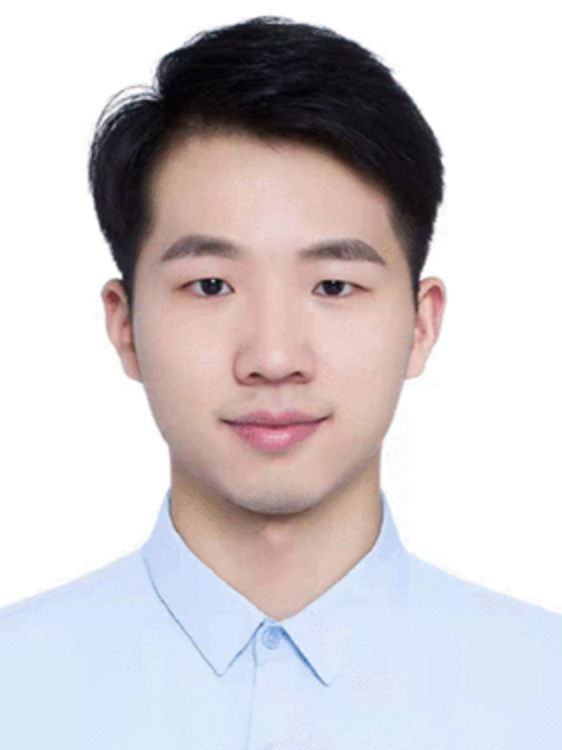}}]{Jinxing Zhou}
received the B.E. and Ph.D. degrees from the School of Computer Science and Information Engineering, Hefei University of Technology, China. He is currently a Postdoctoral Researcher with the Department of Computer Vision, Mohamed bin Zayed University of Artificial Intelligence. His research interests include computer vision, audio-visual learning, and multimedia content analysis.
\end{IEEEbiography}

\vspace{-11mm}

\begin{IEEEbiography}[{\includegraphics[width=1in,height=1.25in,clip,keepaspectratio]{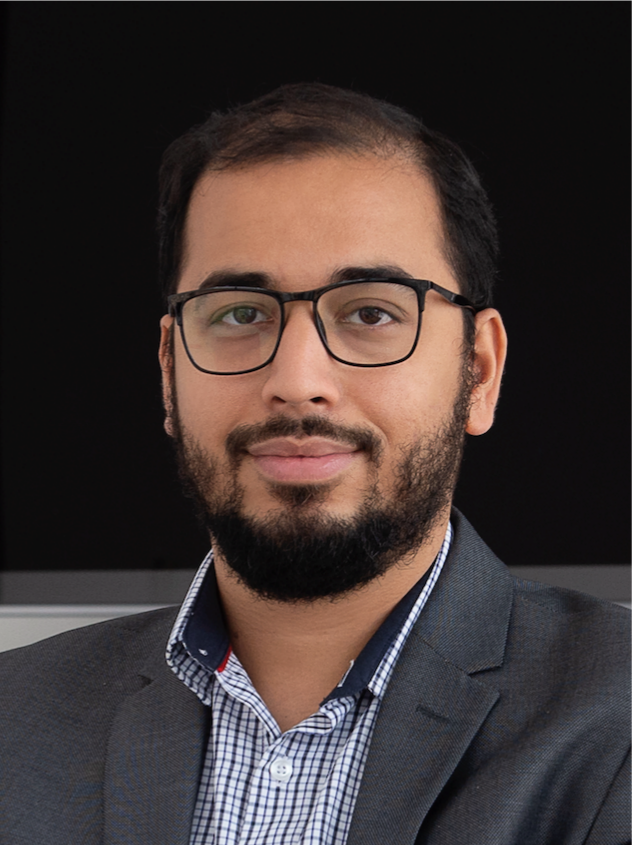}}]{Salman Khan}
(Senior Member, IEEE) received the PhD degree from the University of Western Australia, with a Dean’s List honorable mention. He has been honorary faculty with Australian National University since 2016 and is an Associate Professor with MBZUAI. He serves as Program Chair for ACCV 2028 and has been an Area Chair for CVPR, ICCV, NeurIPS, ICML, ECCV, and ICLR. He is a Guest Editor for IEEE TPAMI. His research interests include computer vision and machine learning.
\end{IEEEbiography}

\vspace{-11mm}

\begin{IEEEbiography}[{\includegraphics[width=1in,height=1.25in,clip,keepaspectratio]{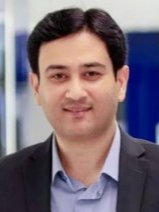}}]{Fahad Shahbaz Khan}
(Senior Member, IEEE) received the PhD degree in computer vision from the Autonomous University of Barcelona, Spain. He is a full professor and deputy department chair of computer vision with MBZUAI, UAE, and faculty member (Universitetslektor + Docent) at Linköping University, Sweden. He serves on the senior program committees of CVPR, ICCV, ECCV, and NeurIPS, will serve as General Chair for ACCV 2028, and is an Associate Editor of IEEE TPAMI and CVIU.
\end{IEEEbiography}

\vfill

\end{document}